\documentclass[]{interact}

\usepackage[caption=false]{subfig}

\usepackage[natbibapa,nodoi]{apacite}
\theoremstyle{plain}

\theoremstyle{definition}

\theoremstyle{remark}

\usepackage{mathtools}
\usepackage{dsfont}
\usepackage{blkarray}
\usepackage{nicematrix}
\usepackage{algorithm}
\usepackage[noend]{algpseudocode}
\algnewcommand\algorithmicinput{\textbf{Input:}}
\algnewcommand\algorithmicoutput{\textbf{Output:}}
\algnewcommand\Input{\item[\algorithmicinput]}%
\algnewcommand\Output{\item[\algorithmicoutput]}%
\algnewcommand{\LineComment}[1]{\State \(\triangleright\) #1}

\usepackage{graphicx}
\usepackage[font={sf,scriptsize},
            labelfont={bf},
            justification= raggedright, 
            singlelinecheck=false]
            {caption}
\begin{document}

\articletype{RESEARCH ARTICLE}

\title{Automation of reversible steganographic coding with\\nonlinear discrete optimisation}

\author{
\name{Ching-Chun Chang\thanks{CONTACT C.-C. Chang. Email: c.c.chang@warwickgrad.net}}
\affil{Department of Computer Science, University of Warwick, Coventry, UK}
}

\maketitle

\begin{abstract}
Authentication mechanisms are at the forefront of defending the world from various types of cybercrime. Steganography can serve as an authentication solution through the use of a digital signature embedded in a carrier object to ensure the integrity of the object and simultaneously lighten the burden of metadata management. Nevertheless, despite being generally imperceptible to human sensory systems, any degree of steganographic distortion might be inadmissible in fidelity-sensitive situations such as forensic science, legal proceedings, medical diagnosis and military reconnaissance. This has led to the development of reversible steganography. A fundamental element of reversible steganography is predictive analytics, for which powerful neural network models have been effectively deployed. Another core element is reversible steganographic coding. Contemporary coding is based primarily on heuristics, which offers a shortcut towards sufficient, but not necessarily optimal, capacity\textendash distortion performance. While attempts have been made to realise automatic coding with neural networks, perfect reversibility is unattainable via such learning machinery. Instead of relying on heuristics and machine learning, we aim to derive optimal coding by means of mathematical optimisation. In this study, we formulate reversible steganographic coding as a nonlinear discrete optimisation problem with a logarithmic capacity constraint and a quadratic distortion objective. Linearisation techniques are developed to enable iterative mixed-integer linear programming. Experimental results validate the near-optimality of the proposed optimisation algorithm when benchmarked against a brute-force method.
\end{abstract}


\section{Introduction}
\label{sec:intro}
Steganography is the art and science of concealing information within a carrier object~\citep{668971}. The term encompasses a wide range of techniques and applications, including but not limited to covert communications~\citep{2005_1511007}, ownership identification~\citep{1997_650120}, copyright protection~\citep{BARNI1998357}, broadcast monitoring~\citep{1999_VIVA} and traitor tracing~\citep{2006_1634364}. An important application of steganography is data authentication, which plays a vital role in cybersecurity. The advent of data-centric artificial intelligence has been accompanied by cybersecurity concerns~\citep{doi:10.1080/09540091.2016.1271400}. It has been reported that intelligent systems are vulnerable to adversarial attacks such as poisonous data collected for re-training during deployment~\citep{Poisoning17}, malware codes hidden in neural network parameters~\citep{10.1145/3427228.3427268} and invisible perturbations crafted to cause erroneous decisions~\citep{2015_Perturb_Goodfellow}. A proper authentication mechanism must ensure that the integrity of data has not been undermined and that the identity of users has not been forged, and thereby protect against these insidious threats.

Digital signatures are a type of authentication technology that is based upon modern cryptography~\citep{10.1145/359340.359342}. This technology can be incorporated into a trustworthy surveillance camera in such a way that photographs are taken and stored along with digital signatures~\citep{267415}. However, storing such auxiliary metadata as a separate file entails the risk of accidental loss and mismanagement during the data lifecycle. Steganography can allow auxiliary information about the data to be embedded invisibly within the data itself. Nevertheless, although generally imperceptible to human sensory systems, any degree of steganographic distortion might not be admissible in some fidelity-sensitive situations such as forensic science, legal proceedings, medical diagnosis and military reconnaissance. This is where the notion of reversible computing comes into play~\citep{2001_Fridrich_Invertible, 2003_1196739, 2004_1315703, 2007_4291553, 2013_6329433, 2018_8534338, ELECTRONIC_IMAGING_2020_Wu}.

A fundamental element of reversible steganography, in common with lossless compression, is predictive modelling~\citep{1948_6773024, 1056936, 623176}. Prediction error modulation is a cutting-edge reversible steganographic technique composed of a \emph{predictive analytics} module and a \emph{reversible coding} module~\citep{2005_1381493, 2007_4099409, 2008Fallahpour, 2009_4811982, 2011_5762603, 2014_6746082, Hwang:2016aa}. The recent development of deep learning has advanced the frontier of reversible steganography. It has been shown that deep neural networks can be applied as powerful predictive models~\citep{2020_9245471, Chang:2021aa, 9736990, Hu:2021aa}. Despite inspiring progress in the analytics module, the design of the coding module is still based largely on heuristics. While there are studies on \emph{end-to-end} deep learning that use neural networks for automatic reversible computing, perfect reversibility cannot be guaranteed~\citep{Jung:2019aa, Duan:2019aa, Lu_2021_CVPR}. From a certain point of view, it is hard for a neural network, as a monolithic black box, to follow the intricate procedures of reversible computing~\citep{Castelvecchi:2016aa}. While deep learning is adept at handling the complex nature of the real world~\citep{LeCun:2015aa}, reversible computing is more of a mechanical process in which procedures have to be conducted in accordance with rigorous algorithms. Therefore, at the time of writing, it seems advisable to follow a \emph{modular} framework.

The essence of reversible steganographic coding is determining how values change to represent different message digits. Different solutions can lead to different trade-offs between capacity and distortion. Instead of relying on heuristics, this study pursues the development of optimal coding for reversible steganography in order to attain optimal capacity\textendash distortion performance. We model reversible steganographic coding as a mathematical optimisation problem and propose an optimisation algorithm for addressing the nonlinearity of this problem. In particular, the task is to minimise steganographic distortion subject to a capacity constraint, where both objective and constraint are nonlinear functions. We propose linearisation techniques for addressing this nonlinear discrete optimisation problem. The remainder of this paper is organised as follows. Section~\ref{sec:back} outlines the background regarding reversible steganography. Section~\ref{sec:optim} formulates the nonlinear discrete optimisation problem and discusses the complexity of a brute-force search algorithm. Section~\ref{sec:linear} presents linearisation techniques for tackling the nonlinear discrete optimisation problem. Section~\ref{sec:sim} analyses the optimality of solutions through simulation experiments. Section~\ref{sec:conclusion} provides concluding remarks.

\begin{figure}[h]
\centerline{\includegraphics[width=0.65\columnwidth]{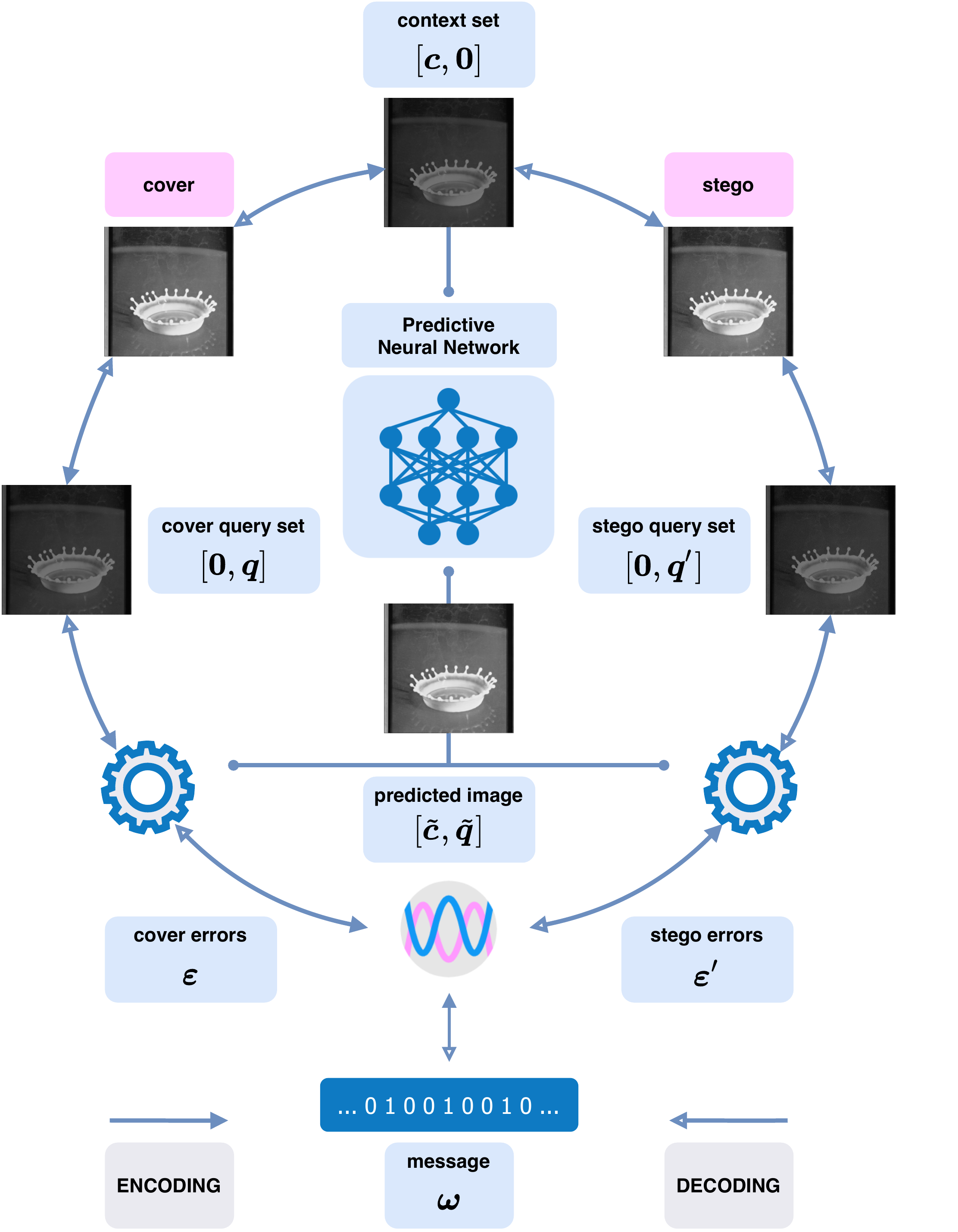}}
\caption{Workflow of reversible steganography with prediction error modulation.}
\label{fig:sys}
\end{figure}

\section{Background}\label{sec:back}
Prediction error modulation is a reversible steganographic technique that consists of an analytics module and a coding module. The analytics module begins by splitting a cover image into \emph{context} and \emph{query} sets, denoted by $\boldsymbol{c}$ and $\boldsymbol{q}$ respectively. A conventional method is to arrange pixels in two groups according to a chequered pattern. Then a predictive model is applied to predict the intensities of the query pixels from the intensities of the context pixels. A contemporary practice of predictive modelling is to employ an artificial neural network originally designed for computer vision tasks. The coding module embeds a message $\boldsymbol{\omega}$ into the cover image by modulating the prediction errors $\boldsymbol{\varepsilon} = \boldsymbol{q} - \tilde{\boldsymbol{q}}$. The modulated errors $\boldsymbol{\varepsilon}^{\prime}$ are then added to the predicted intensities, causing distortion of the query pixels. The stego image is created by merging the context set $\boldsymbol{c}$ and the modulated query set $\boldsymbol{q}^{\prime}$. The decoding procedure is similar to the encoding procedure. It begins by predicting the query pixel intensities. Since the context set is kept unchanged, the prediction in the decoding phase is guaranteed to be identical to that in the encoding phase given the same predictive model. The message is extracted and the query set is recovered by demodulating the prediction errors. The image is reversed to its original state by merging the context and recovered query sets. The procedures for encoding and decoding are depicted schematically in Figure~\ref{fig:sys} and also provided in Algorithms~\ref{alg:enc} and~\ref{alg:dec}. We would like to note that the message may contain certain auxiliary information for handling pixel intensity overflow. This paper does not go into detail about every aspect of the stego-system; instead, our study focuses on the mathematical optimisation of reversible steganographic coding.

\begin{minipage}{0.45\textwidth}
\begin{algorithm}[H]
\centering
\caption{Encoding}\label{alg:enc}
\begin{algorithmic}

\Input $\text{cover}$, $\boldsymbol{\omega}$
\Output $\text{stego}$
\\
\LineComment{analytics module}
\State $[\boldsymbol{c}, \boldsymbol{q}] \gets \operatorname{split}(\text{cover})$
\State $[\tilde{\boldsymbol{c}}, \tilde{\boldsymbol{q}}] \gets \operatorname{predict}([\boldsymbol{c},\boldsymbol{0}])$
\\
\LineComment{coding module}
\State $\boldsymbol{\varepsilon} \gets \boldsymbol{q} - \tilde{\boldsymbol{q}}$
\State $\boldsymbol{\varepsilon}^{\prime} \gets \operatorname{modulate}(\boldsymbol{\varepsilon}, \boldsymbol{\omega})$
\State $\boldsymbol{q}^{\prime} \gets \tilde{\boldsymbol{q}} + \boldsymbol{\varepsilon}^{\prime}$
\State $\text{stego} \gets \operatorname{merge}(\boldsymbol{c}, \boldsymbol{q}^{\prime})$
\end{algorithmic}
\end{algorithm}
\end{minipage}
\hspace{5pt}
\begin{minipage}{0.45\textwidth}
\begin{algorithm}[H]
\centering
\caption{Decoding}\label{alg:dec}
\begin{algorithmic}

\Input $\text{stego}$
\Output $\text{cover}$, $\boldsymbol{\omega}$
\\
\LineComment{analytics module}
\State $[\boldsymbol{c}, \boldsymbol{q}^{\prime}] = \operatorname{split}(\text{stego})$
\State $[\tilde{\boldsymbol{c}}, \tilde{\boldsymbol{q}}] = \operatorname{predict}([\boldsymbol{c}, \boldsymbol{0}])$
\\
\LineComment{coding module}
\State $\boldsymbol{\varepsilon}^{\prime} = \boldsymbol{q}^{\prime} - \tilde{\boldsymbol{q}}$
\State $[\boldsymbol{\varepsilon}, \boldsymbol{\omega}] = \operatorname{demodulate}(\boldsymbol{\varepsilon}^{\prime})$
\State $\boldsymbol{q} = \tilde{\boldsymbol{q}} + \boldsymbol{\varepsilon}$
\State $\text{cover} = \operatorname{merge}(\boldsymbol{c}, \boldsymbol{q})$
\end{algorithmic}
\end{algorithm}
\end{minipage}

\section{Nonlinear Discrete Optimisation}\label{sec:optim}
The essence of reversible steganographic coding is designating one or multiple error values as the carrier and determining how these values change to represent different message digits. A rule of thumb for reversible steganographic coding is to choose the prediction errors of the peak frequency as the carrier. While the peak frequency implies the highest capacity, this capacity-greedy strategy is not necessarily optimal in terms of minimising distortion.

\subsection{Problem Definition}
\begin{figure}
\centerline{\includegraphics[width=0.65\columnwidth]{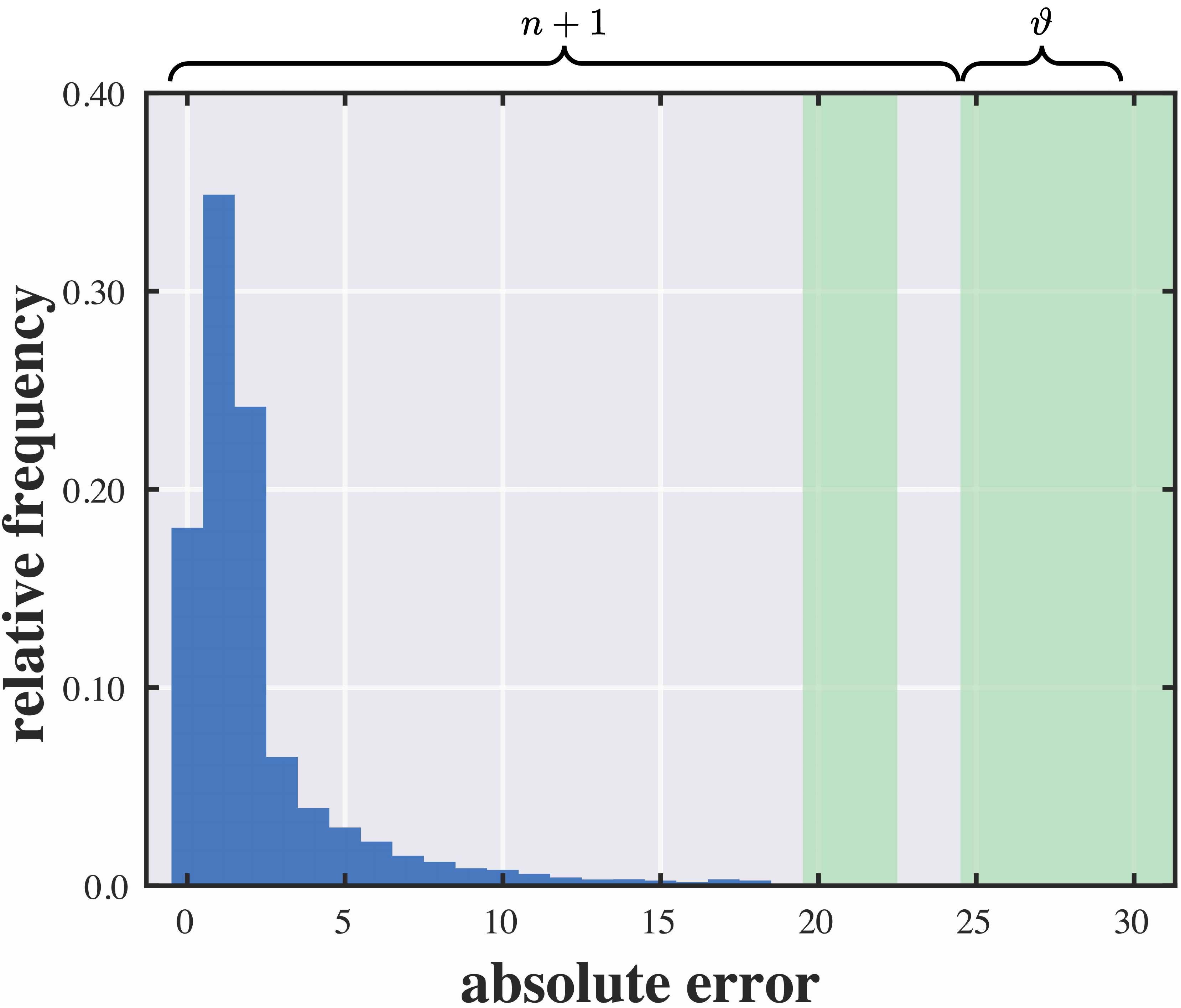}}
\caption{Example of absolute error distribution with highlighted zero occurrences.}
\label{fig:error_distrib}
\end{figure}

\begin{figure}
\centerline{\includegraphics[width=0.88\columnwidth]{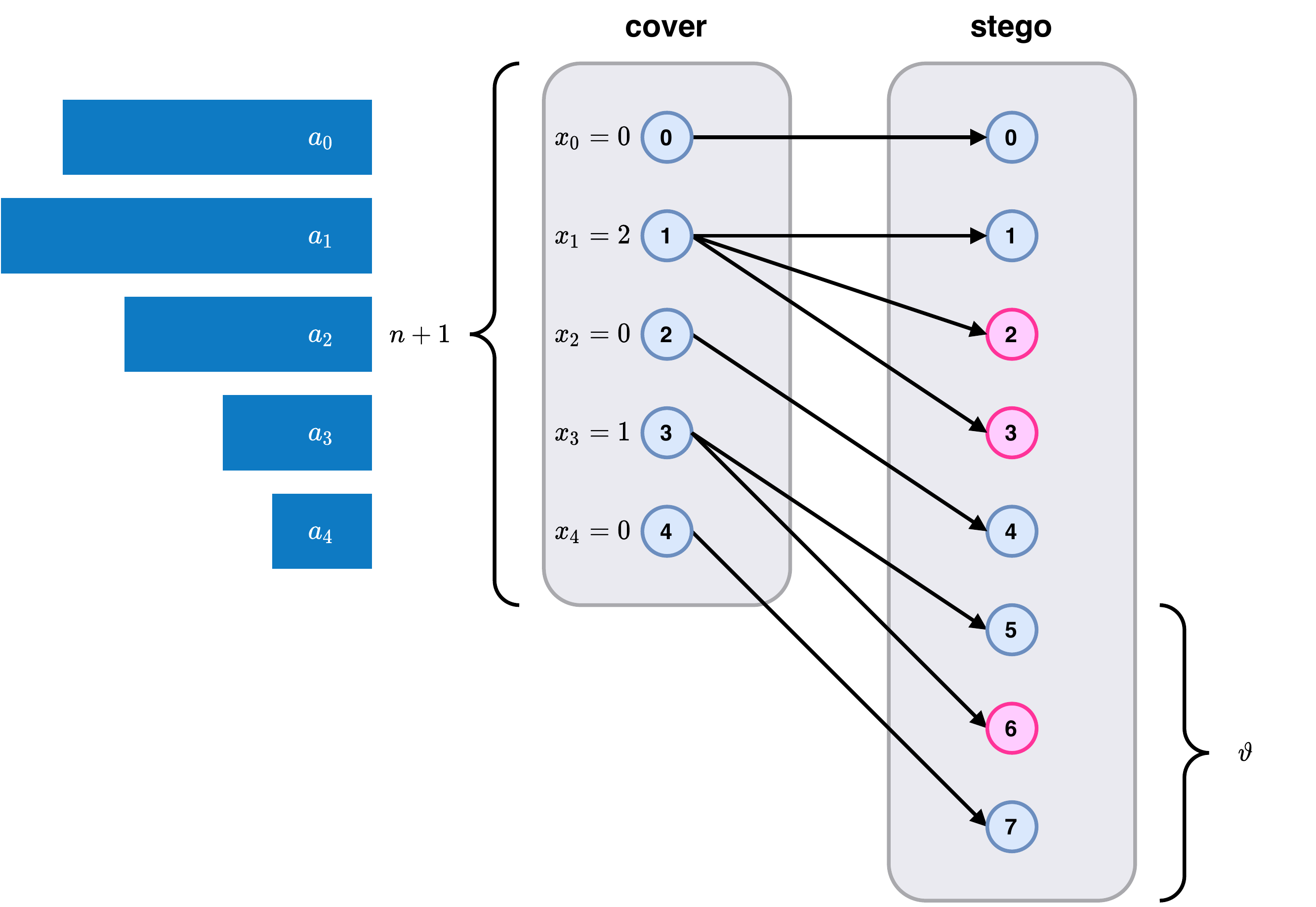}}
\caption{Example of reversible steganographic coding.}
\label{fig:mapping}
\end{figure}

According to the typical law of error, the frequency of an error can be expressed as an exponential function of its numerical magnitude, disregarding sign~\citep{Wilson:1923aa}. In other words, the frequency distribution of prediction errors is expected to centre around zero. In general, a smaller absolute error tends to have a higher occurrence. A special exception is that the occurrence of zero might be lower than the occurrence of a certain small absolute error considering that the latter is the sum of both positive and negative error occurrences. Consider an absolute error histogram as shown in Figure~\ref{fig:error_distrib}. The problem of reversible steganographic coding is to establish a mapping between the values in $[0, n]$ and the values in $[0, n+\vartheta]$, where $\vartheta$ denotes the extra quota and is typically defined as less than or equal to the number of successive empty bins in the absolute error histogram. Encoding is a \emph{one-to-many} mapping that links a cover value to one or more stego values. A message digit can only be represented if the connections are greater than one. Different cover values can never yield the same stego value. This is done in order to avoid an overlap between values (i.e. an ambiguity in decoding). Therefore, a cover value may be changed to a different stego value even if it does not represent any message digit. We impose a constraint that each cover value can only be mapped to the nearest available stego values since a \emph{non-cross} mapping drastically reduces the problem dimension. An example of a cover/stego mapping is illustrated in Figure~\ref{fig:mapping}. 

\subsection{Model Formulation}
Let us denote by $a_i$ the frequency of the value $i$ and by $x_i$ the number of extra cover-to-stego links for the value $i$. The total number of links for the value $i$ equals $x_i + 1$. The number of bits that can be represented by modifying the value $i$ is $\log_2(x_i + 1)$ and thus the capacity is computed by
\begin{equation}
\mathfrak{c} = \sum_{i=0}^n a_i \log_2(x_i + 1)	.
\end{equation}
In fact, the number of bits that can be represented by modifying $0$ equals $\log_2 (2x_0+1)$ because $0$ can be mapped to both positive and negative values. For example, there are three different states $0$ and $\pm1$ when $x_0=1$, and five different states $0$, $\pm1$ and $\pm2$ when $x_0=2$. To be concise, we simplify the case by mapping $0$ randomly to a positive or negative value so that the capacity computation for $0$ is identical to that for other values at the cost of slightly underestimating the capacity offered by the former. The probability of changing a cover value to each stego value is $1/(x_i + 1)$. The deviations of the first to the last stego value are $0 + y_i$ to $x_i + y_i$ respectively, where $y_i$ denotes the sum of all the previous extra links (i.e. the cumulative deviation). Hence, the expected distortion in terms of the squared deviations is computed by
\begin{equation}
\mathfrak{D} = \sum_{i=0}^n a_i \left(\frac{(0 + y_i)^2 + (1 + y_i)^2 + \dots + (x_i + y_i)^2}{x_i + 1}\right),
\end{equation}
where
\begin{equation}
y_i = \sum_{j=0}^{i-1} x_j .
\end{equation}
We can simplify the algebraic expression by
\begin{equation}
\begin{split}
&\frac{(0+ y_i)^2 + (1+ y_i)^2 + \dots + (x_i + y_i)^2}{x_i + 1} \\
= &\frac{ (0^2 + 2y_i\cdot 0 + y_i^2) + (1^2 + 2y_i\cdot 1 + y_i^2) + \dots +(x_i^2 + 2y_i\cdot x_i + y_i^2) }{x_i + 1} \\
= &\frac{ (0^2 + 1^2 + \dots + x_i^2) + 2y_i(0 + 1 + \dots + x_i) + y_i^2(x_i+1) }{x_i + 1}\\
= &\frac{x_i(x_i+1)(2x_i+1)}{6(x_i+1)} + \frac{2y_ix_i(x_i+1)}{2(x_i+1)} + \frac{y_i^2(x_i+1)}{x_i+1}\\
= &\frac{1}{3}x_i^2 + \frac{1}{6}x_i + x_iy_i + y_i^2 .\\
\end{split}
\end{equation}
The reason for computing squared deviations rather than absolute deviations is that image quality is often measured by the peak signal-to-noise ratio (PSNR), which is defined via the mean squared error (MSE). Our goal is to solve for the decision variables $x_i \in \{0,\dots \vartheta\}$ which minimise the distortion objective subject to the capacity constraint. The sum of all the extra cover-to-stego links is not allowed to exceed the quota $\vartheta$. To summarise, the mathematical optimisation problem for reversible steganographic coding is
\begin{equation*}
\begin{alignedat}{3}
& \text{min} & \enspace & \mathfrak{D} = \sum_{i=0}^n a_i \left( \frac{1}{3}x_i^2 + \frac{1}{6}x_i + x_iy_i + y_i^2 \right) ,\\
& \text{s.t.} & & \mathfrak{c} = \sum_{i=0}^n a_i \log_2(x_i + 1)	 \geq \text{payload} ,\\
&&& \sum_{i=0}^n x_i \leq \vartheta ,\\
& \text{var.} && x_i \in \{0,\cdots,\vartheta\}, \quad & \hspace{-0.5cm} \forall i=0,\dots,n .
\end{alignedat}
\end{equation*}

\subsection{Brute-Force Search}
Brute-force search is a baseline method for benchmarking optimisation algorithms. The solution space that exhausts all possible combinations of the decision variables is equal to $(\vartheta+1)^{n+1} \in \mathcal{O}(c^n)$. By taking account of the quota constraint, we can reduce the solution space from the number of possible combinations to the number of feasible combinations. In number theory and combinatorics, the partition function $\operatorname{part}(t)$ computes the number of ways of writing $t$ as a sum of the positive integers in $[1,t]$. Let $\boldsymbol{\Lambda}_{t}$ denote a matrix of $\operatorname{part}(t)$ rows and $t$ columns which enumerates all possible partitions:
\begin{equation}
\boldsymbol{\Lambda}_{t} =
\begin{bNiceArray}{*{1}{c}}
\boldsymbol{\lambda}_{1}\\
\vdots\\
\boldsymbol{\lambda}_{\operatorname{part}(t)}\\
\end{bNiceArray}
=
\begin{bNiceArray}{*{3}{c}}[]
    \lambda_{1,1} & \cdots & \lambda_{1,t}\\
    \vdots     & \ddots & \vdots 		   \\
    \lambda_{\operatorname{part}(t),1} & \cdots & \lambda_{\operatorname{part}(t),t} \\
\end{bNiceArray}.	
\end{equation}
Each vector $\boldsymbol{\lambda}_{\ell}$ represents a possible partition in which each element is the quantity of a candidate integer (i.e. the summand). For example, $\boldsymbol{\Lambda}_{2}$, $\boldsymbol{\Lambda}_{3}$ and $\boldsymbol{\Lambda}_{4}$ are
\begin{equation*}
\begin{bNiceArray}{*{2}{c}}[first-row,last-col,code-for-first-row=\scriptscriptstyle,code-for-last-col=\scriptscriptstyle]
    1 & 2 \\
    2 & 0 & \boldsymbol{\lambda}_{1} \\
    0 & 1 & \boldsymbol{\lambda}_{2} \\
\end{bNiceArray},\quad
\begin{bNiceArray}{*{3}{c}}[first-row,last-col,code-for-first-row=\scriptscriptstyle,code-for-last-col=\scriptscriptstyle]
    1 & 2 & 3 & \\
    3 & 0 & 0 & \boldsymbol{\lambda}_{1} \\
    1 & 1 & 0 & \boldsymbol{\lambda_{2}} \\
    0 & 0 & 1 & \boldsymbol{\lambda_{3}} \\
\end{bNiceArray},\quad
\begin{bNiceArray}{*{4}{c}}[first-row,last-col,code-for-first-row=\scriptscriptstyle,code-for-last-col=\scriptscriptstyle]
    1 & 2 & 3 & 4 & \\
    4 & 0 & 0 & 0 & \boldsymbol{\lambda_{1}} \\
    2 & 1 & 0 & 0 & \boldsymbol{\lambda_{2}} \\
    0 & 2 & 0 & 0 & \boldsymbol{\lambda_{3}} \\
    1 & 0 & 1 & 0 & \boldsymbol{\lambda_{4}} \\
    0 & 0 & 0 & 1 & \boldsymbol{\lambda_{5}} \\
\end{bNiceArray}.	
\end{equation*}
The total number of feasible solutions can be calculated by adding up the number of feasible solutions given by each individual partition matrix from $\boldsymbol{\Lambda}_{1}$ to $\boldsymbol{\Lambda}_{\vartheta}$; that is,
\begin{equation}
\sum_{t=1}^{\vartheta} \operatorname{feasible}(\boldsymbol{\Lambda}_{t}, n^*) ,
\end{equation}
where $n^* = n + 1$ denotes the number of integers in $[0, n]$. For each matrix $\boldsymbol{\Lambda}_{t}$, the number of feasible solutions is computed by summing the number of possible combinations given by each partition vector $\boldsymbol{\lambda}_{\ell}$, denoted by
\begin{equation}
\operatorname{feasible}(\boldsymbol{\Lambda}_{t}, n^*) = \sum_{\ell=1}^{\operatorname{part}(t)} \operatorname{comb}(\boldsymbol{\lambda}_{\ell}, n^*) .
\end{equation}
A combination is a selection of values from a set of $n^*$ values based on a given partition vector and hence the number of combinations is computed by
\begin{equation}
\operatorname{comb}(\boldsymbol{\lambda}_{\ell}, n^*) = \prod_{i=1}^{t} \binom{n^* - \sum_{j=1}^{i-1} \lambda_{j}^* }{\lambda_{i}^*} ,
\end{equation}
where $\lambda_{i}^* = \lambda_{\ell,i}$ represents a convenient notation without explicitly writing out the index of the partition vector (for reducing the verbosity). The number of combinations is a product of $t$ binomial coefficients and each term is meant to choose (and remove) an unordered subset of $\lambda_{i}^*$ values from the remaining values in the set of $n^*$ values. Let us take $\boldsymbol{\Lambda}_{3}$ for example. The number of combinations for partition vectors $\boldsymbol{\lambda}_{1}$, $\boldsymbol{\lambda}_{2}$ and $\boldsymbol{\lambda}_{3}$ are computed as follows:
\begin{equation*}
\operatorname{comb}(\boldsymbol{\lambda}_{1}, n^*) = \binom{n^*}{3}\binom{n^* - 3}{0}\binom{n^* - 3 - 0}{0} ,
\end{equation*}

\begin{equation*}
\operatorname{comb}(\boldsymbol{\lambda}_{2}, n^*) = \binom{n^*}{1}\binom{n^* - 1}{1}\binom{n^* - 1 - 1}{0} ,
\end{equation*}

\begin{equation*}
\operatorname{comb}(\boldsymbol{\lambda}_{3}, n^*) = \binom{n^*}{0}\binom{n^* - 0}{0}\binom{n^* - 0 - 0}{1} .
\end{equation*}
The number of combinations can be approximated by
\begin{equation}
\begin{split}
& \prod_{i=1}^{t} \binom{n^* - \sum_{j=1}^{i-1} \lambda_{j}^* }{\lambda_{i}^*} \\
= & \binom{n^*}{\lambda_1^*} \binom{n^* - \lambda_1^*}{\lambda_2^*} \cdots \binom{n^* - \lambda_1^* - \lambda_2^* - \dots - \lambda_{t-1}^* }{\lambda_{t}^*}\\
= & \frac{n^*!}{\lambda_1^*!(n^*-\lambda_1^*)!} \times \frac{(n^*-\lambda_1^*)!}{\lambda_2^*!(n^*- \lambda_1^*-\lambda_2^*)!} \times \dots \times \frac{(n-\sum_{j=1}^{t-1} \lambda_{j}^*)!}{\lambda_{t}^*!(n-\sum_{j=1}^{t} \lambda_{j}^*)!}\\
= & \frac{n^*!}{\lambda_1^*!\lambda_2^*!\dots \lambda_{t}^*!(n^*-\sum_{j=1}^{t} \lambda_{j}^*)!}\\
= & \frac{n^*(n^*-1)(n^*-2)\dots(n^*-(\sum_{j=1}^{t} \lambda_{j}^* - 1))}{\lambda_1^*!\lambda_2^*!\dots \lambda_{t}^*!}\\
\leq & \frac{n^*(n^*-1)(n^*-2)\dots(n^*- (t - 1))}{\lambda_1^*!\lambda_2^*!\dots \lambda_{t}^*!} 
\approx n^{t} .
\end{split}
\end{equation}
Hence, the complexity of this brute-force algorithm is approximately equal to
\begin{equation}
\sum_{t=1}^{\vartheta} \sum_{\ell=1}^{\operatorname{part}(t)} \operatorname{comb}(\boldsymbol{\lambda}_{\ell}, n^*) \approx \sum_{t=1}^{\vartheta} \operatorname{part}(t)\cdot n^{t} \in \mathcal{O}(n^c) .
\end{equation}

\begin{figure}
\centerline{\includegraphics[width=0.7\columnwidth]{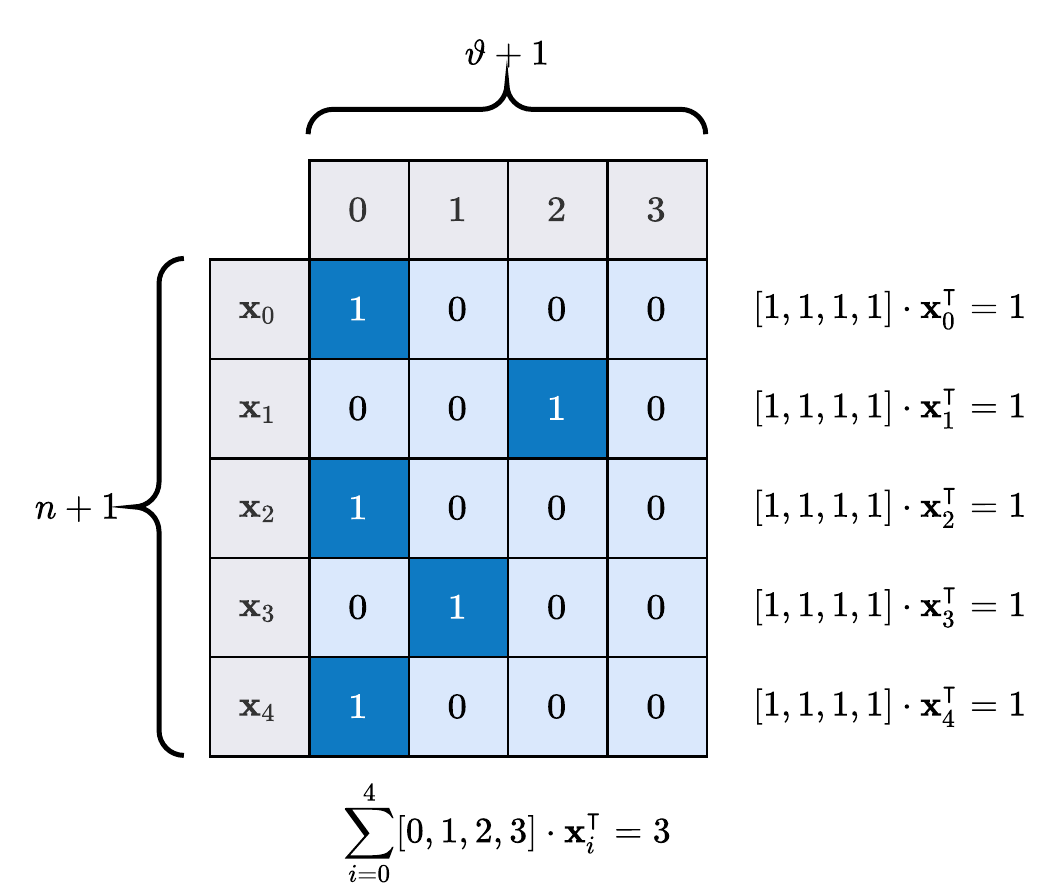}}
\caption{Example of binary-integer decision variable.}
\label{fig:onehot}
\end{figure}

\section{Linearisation}\label{sec:linear}
The difficulty of our optimisation problem lies in the nonlinear nature of the capacity constraint and the distortion objective. To apply off-the-shelf optimisation tools, we have to tackle these nonlinearities.

\subsection{Logarithmic Capacity Constraint}
The capacity constraint involves the calculation of logarithm of variables $\log_2(x_i+1)$. The logarithmic function is nonlinear. A useful linearisation trick is to re-model the problem with binary-integer variables. We binarise each decision variable $x_i$ with the domain $[0, \vartheta]$ into a 0/1 vector (or a one-hot vector) of length $\vartheta + 1$, as illustrated in Figure~\ref{fig:onehot}. The vector consists of 0s with the exception of a single 1 whose position indicates the value of $x_i$; that is,
\begin{equation}
\mathbf{x}_i = [\mathrm{x}_i^0, \cdots, \mathrm{x}_i^{\vartheta}] \in \{0,1\}^{\vartheta + 1}	,
\end{equation}
such that
\begin{equation}
\mathds{1} \cdot \mathbf{x}_i^{\intercal} = 1, \quad \forall i = 0, \dots ,n .
\end{equation}
We can retrieve $x_i$ using the dot product of vectors
\begin{equation}
x_i = [0,\cdots,\vartheta] \cdot \mathbf{x}_i^{\intercal} = \mathbf{v}\mathbf{x}_i^{\intercal} .
\end{equation}
Accordingly, the quota constraint becomes
\begin{equation}
\sum_{i=0}^n \mathbf{v}\mathbf{x}_i^{\intercal} \leq \vartheta .
\end{equation}
In a similar manner, the logarithm can be derived using the dot product of vectors
\begin{equation}
\begin{split}
\log_2(x_i+1) &= \left[ \log_2(0+1), \cdots, \log_2(\vartheta+1) \right] \cdot \mathbf{x}_i^{\intercal}\\
&= \mathbf{v}_{\operatorname{log}} \mathbf{x}_i^{\intercal} .
\end{split}
\end{equation}
Hence, we rewrite the capacity constraint as
\begin{equation}
\mathfrak{c} = \sum_{i=0}^n a_i \mathbf{v}_{\operatorname{log}} \mathbf{x}_i^{\intercal} .
\end{equation}

\subsection{Quadratic Distortion Objective}
The distortion objective involves three nonlinear terms $x_i^2$, $y_i^2$ and $x_i y_i$. These terms are quadratic functions of variables. The first term can be approached using the dot product as before; that is  
\begin{equation}
x_i^2 = [0^2, \dots, \vartheta^2]\cdot \mathbf{x}_i^{\intercal}
= \mathbf{v}_{\operatorname{sq}} \mathbf{x}_i^{\intercal} .
\end{equation}
The remaining two terms contain the partial sum of variables $y_i$, which is computed by
\begin{equation}
y_i = \sum_{j=0}^{i-1}  \mathbf{v}\mathbf{x}_j^{\intercal} .
\end{equation}
To linearise the univariate quadratic term $y_i^2$ and the bivariate quadratic term $x_i y_i$, we introduce two non-negative continuous \emph{slack} variables $z_{y_i^2} \geq 0$ and $z_{x_iy_i} \geq 0$. Replacing the quadratic terms with the dot product and the slack variables results in a linear distortion objective    
\begin{equation}
\mathfrak{D} = \sum_{i=0}^n a_i \left( \frac{1}{3}\mathbf{v}_{\operatorname{sq}} \mathbf{x}_i^{\intercal} + \frac{1}{6}\mathbf{v} \mathbf{x}_i^{\intercal} + z_{x_iy_i} + z_{y_i^2} \right) .
\end{equation}
We begin by solving this mixed-integer linear programming problem, which does not yet reflect the quadratic terms regarding cumulative distortion, and obtain an initial solution comprising $\tilde{\mathbf{x}}_i$, $\tilde{z}_{y_i^2}$, and $\tilde{z}_{x_iy_i}$. The initial slack variables would be zeros because the objective is to minimise distortion. To make the slack variables reflect the quadratic terms properly, we add the following constraints
\begin{equation}
\begin{alignedat}{2}
z_{y_i^2} &\geq y_i^2,\\
z_{x_iy_i} &\geq x_iy_i.
\end{alignedat}
\end{equation}
In this way, we reformulate a problem with a nonlinear objective into a problem with a linear objective and nonlinear constraints. We make use of the solution obtained previously to linearise these nonlinear constraints and solve the mixed-integer linear programming problem iteratively. To begin with, we express the variables in terms of the previous solution:
\begin{equation}
\begin{split}
x_i &= \tilde{x}_{i} + \delta_{x_i} ,\\
y_i &= \tilde{y}_{i} + \delta_{y_i} ,
\end{split}
\end{equation}
where $\tilde{x}_{i}$ and $\tilde{y}_{i}$ are treated as constants. Then, we apply the first-order Taylor series to approximate the univariate quadratic term as
\begin{equation}
\begin{split}
f(y_i) &= f(\tilde{y}_{i} + \delta_{y_i}) \\
&= f(\tilde{y}_{i}) + f^{\prime}(\tilde{y}_{i}) \delta_{y_i} + \cdots \\
&= \tilde{y}_{i}^2 + 2\tilde{y}_{i}\delta_{y_i} + \cdots \\
&\approx \tilde{y}_{i}^2 +  2\tilde{y}_{i}(y_i - \tilde{y}_{i})\\
&= 2\tilde{y}_{i}y_i - \tilde{y}_{i}^2	,
\end{split}
\end{equation}
and similarly the bivariate quadratic term as
\begin{equation}
\begin{split}
f(x_i, y_i) &= f(\tilde{x}_{i} +\delta_{x_i}, \tilde{y}_{i} + \delta_{y_i})\\
&= f(\tilde{x}_{i},\tilde{y}_{i}) + \frac{\partial f(\tilde{x}_{i},\tilde{y}_{i})}{\partial x_i} \delta_{x_i} + \frac{\partial f(\tilde{x}_{i},\tilde{y}_{i})}{\partial y_i} \delta_{y_i} + \cdots \\
&= \tilde{x}_{i}\tilde{y}_{i} + \tilde{y}_{i}\delta_{x_i} + \tilde{x}_{i}\delta_{y_i} + \cdots \\
&\approx \tilde{x}_{i}\tilde{y}_{i} + \tilde{y}_{i}(x_i - \tilde{x}_{i}) + \tilde{x}_{i}(y_i - \tilde{y}_{i})\\
&= \tilde{x}_{i}y_i + \tilde{y}_{i}x_i - \tilde{x}_{i}\tilde{y}_{i} .
\end{split}
\end{equation}
As a result, the nonlinear constraints are transformed into linear constraints
\begin{equation}
\begin{alignedat}{2}
2\tilde{y}_{i}y_i - z_{y_i^2} &\leq \tilde{y}_{i}^2 ,\\
\tilde{x}_{i}y_i + \tilde{y}_{i}x_i - z_{x_iy_i} &\leq \tilde{x}_{i}\tilde{y}_{i} .
\end{alignedat}
\end{equation}
To recapitulate, the nonlinear discrete optimisation problem is approached by means of an iterative method that solves a mixed-integer linear programming problem with binary-integer variables and non-negative continuous slack variables:
\begin{equation*}
\begin{alignedat}{3}
& \text{min} & \enspace & \mathfrak{D} = \sum_{i=0}^n a_i \left( \frac{1}{3}\mathbf{v}_{\operatorname{sq}} \mathbf{x}_i^{\intercal} + \frac{1}{6}\mathbf{v} \mathbf{x}_i^{\intercal} + z_{x_iy_i} + z_{y_i^2} \right) ,\\
& \text{s.t.} & \enspace & \mathfrak{c} = \sum_{i=0}^n a_i \mathbf{v}_{\operatorname{log}} \mathbf{x}_i^{\intercal} \geq \text{payload} ,\\
&&& \sum_{i=0}^n \mathbf{v}\mathbf{x}_i^{\intercal} \leq \vartheta ,\\
&&& \mathds{1} \cdot \mathbf{x}_i^{\intercal} = 1, \quad & \hspace{-2.0cm} \forall i = 0, \dots ,n ,\\
&&& 2\tilde{y}_{i}y_i - z_{y_i^2} \leq \tilde{y}_{i}^2, \quad & \hspace{-2.0cm} \forall i=0, \dots, n ,\\
&&& \tilde{x}_{i}y_i + \tilde{y}_{i}x_i - z_{x_iy_i} \leq \tilde{x}_{i}\tilde{y}_{i}, \quad &\hspace{-2.0cm} \forall i=0, \dots, n ,\\
& \text{var.} & \enspace & \mathbf{x}_i \in \{0,1\}^{\vartheta + 1}, \quad & \hspace{-2.0cm} \forall i = 0, \dots ,n ,\\
&&& z_{y_i^2} \geq 0, \quad & \hspace{-2.0cm} \forall i = 0, \dots ,n ,\\
&&& z_{x_iy_i}\geq 0, \quad & \hspace{-2.0cm} \forall i = 0, \dots ,n ,\\
& \text{where} & & x_i = \mathbf{v} \mathbf{x}_i^{\intercal} \quad \& \quad y_i = \sum_{j=0}^{i-1}  \mathbf{v}\mathbf{x}_j^{\intercal} .
\end{alignedat}
\end{equation*}

\begin{figure}
\centering
\subfloat[Aeroplane]{%
\resizebox*{5.5cm}{!}{\includegraphics{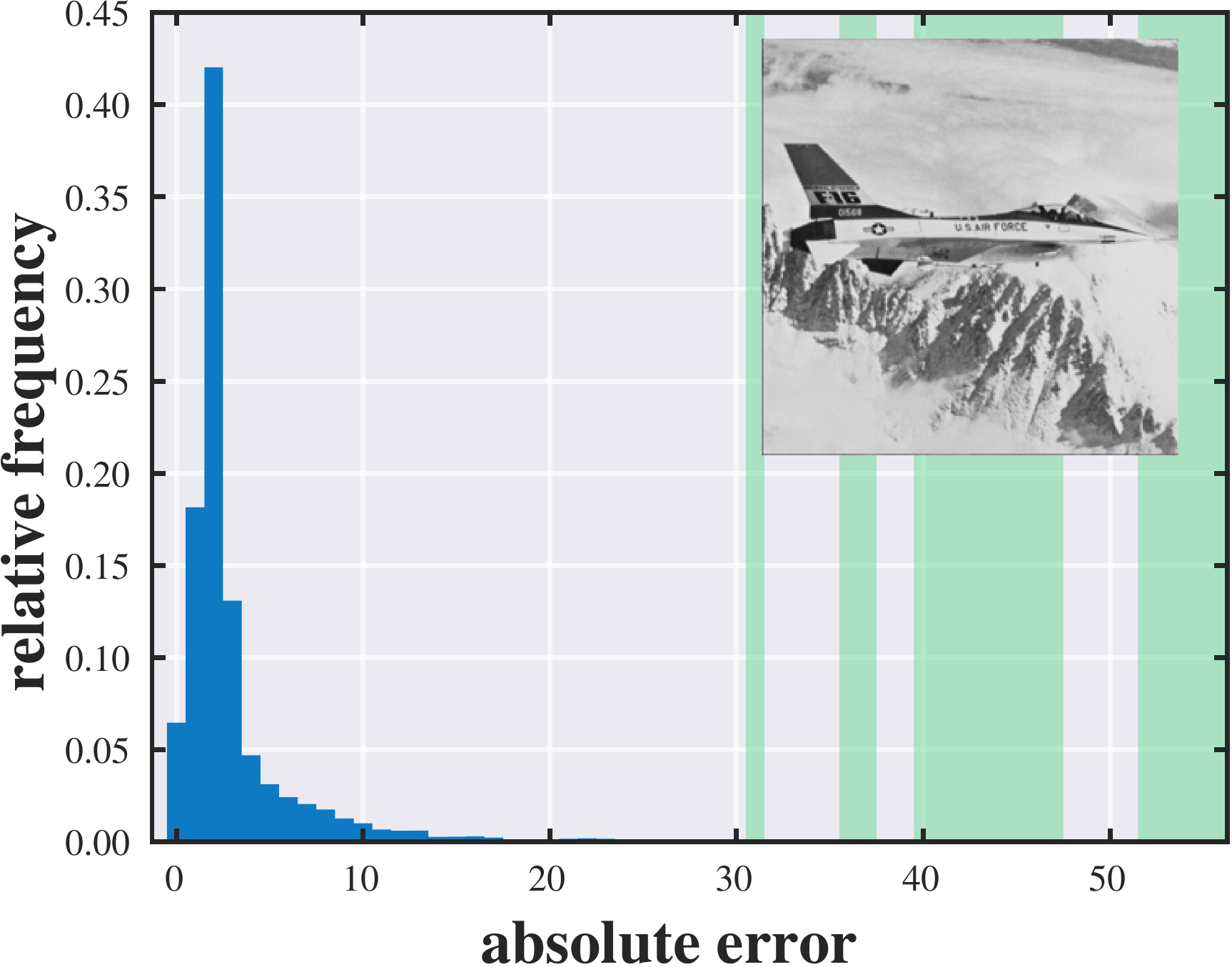}}}\hspace{5pt}
\subfloat[Lena]{%
\resizebox*{5.5cm}{!}{\includegraphics{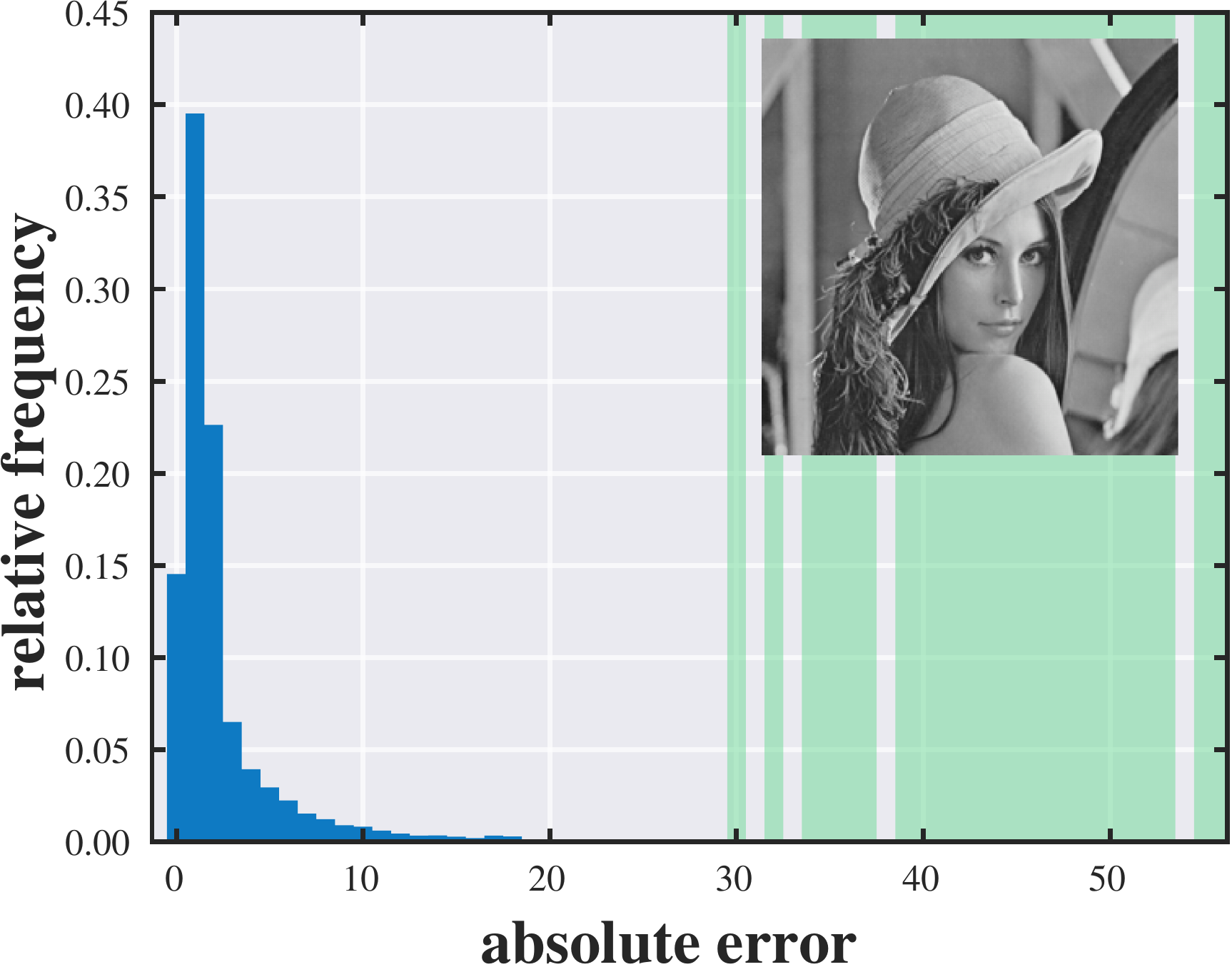}}}
\\
\subfloat[Mandrill]{%
\resizebox*{5.5cm}{!}{\includegraphics{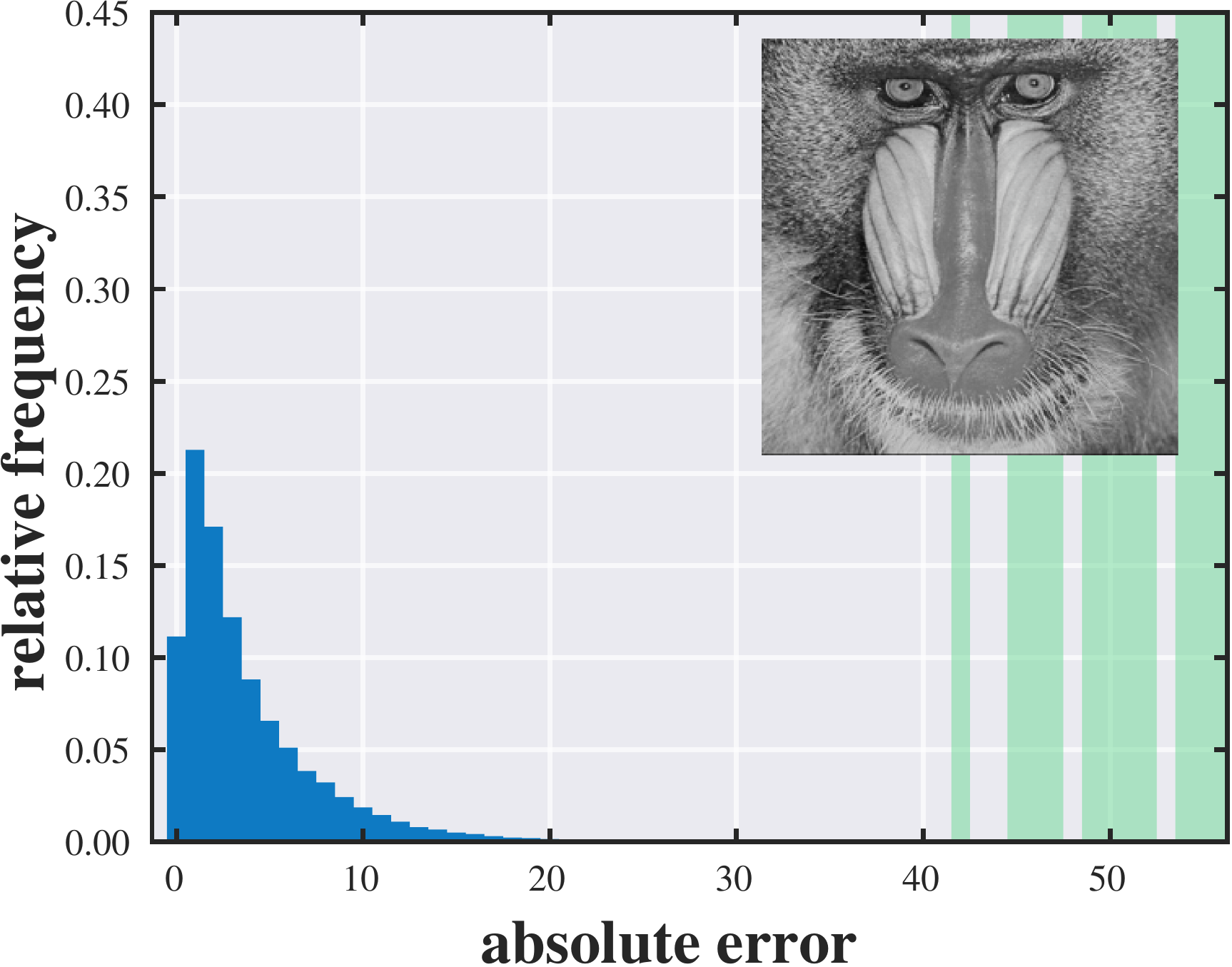}}}\hspace{5pt}
\subfloat[Peppers]{%
\resizebox*{5.5cm}{!}{\includegraphics{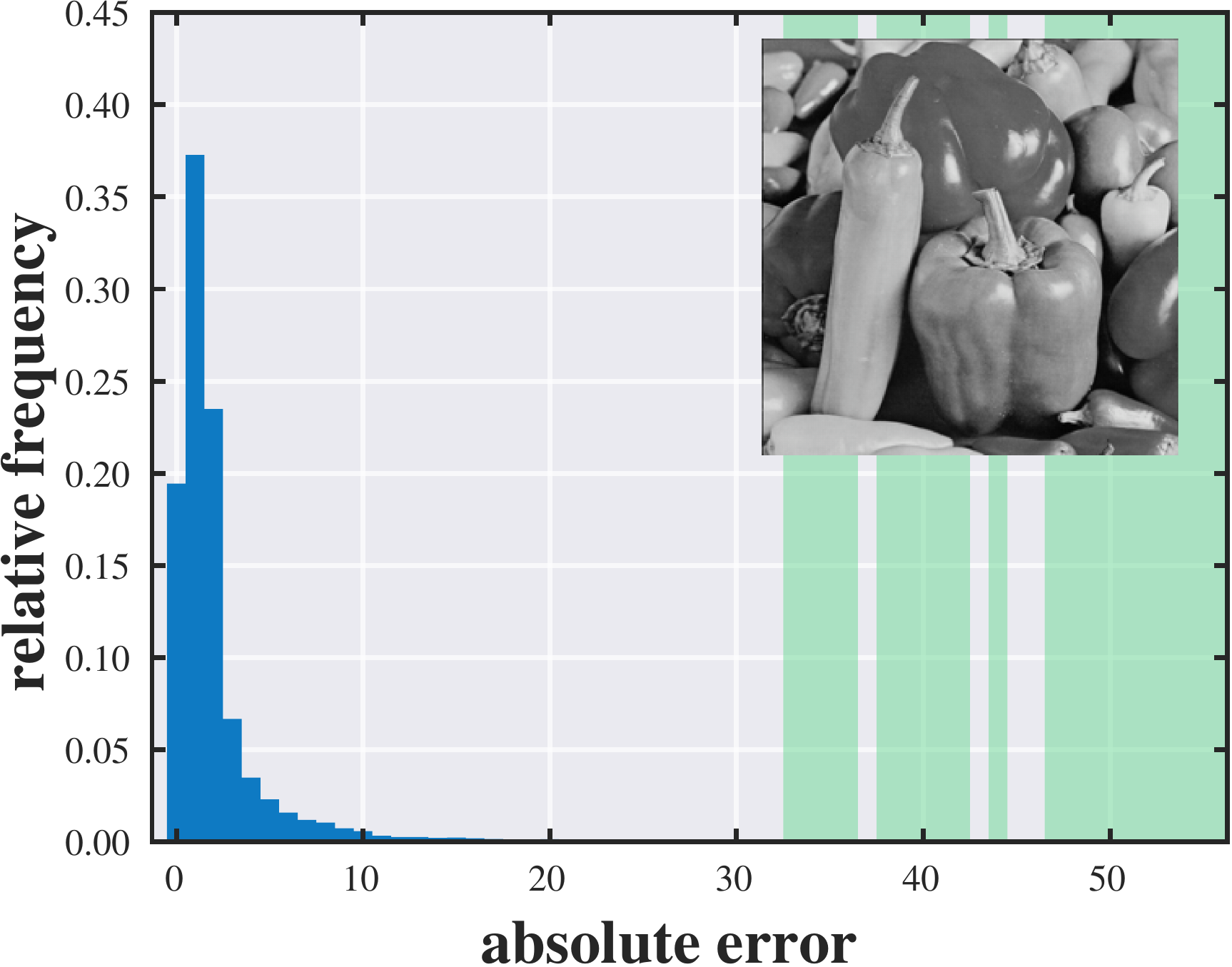}}}
\caption{Absolute error histograms with highlighted empty bins.} \label{fig:img_hist}
\end{figure}

\section{Simulation}\label{sec:sim}
We carry out experimental analysis on the optimality of the proposed method benchmarked against the brute-force method. The experimental setup is described as follows. For the predictive model, we use the residual dense network (RDN), which has its origins in low-level computer vision tasks such as super-resolution imaging~\citep{2018_8578360} and image restoration~\citep{8964437}. This neural network model is characterised by a tangled labyrinth of residual and dense connections. It is trained on the BOSSbase dataset~\citep{2011_BOSSbase}, which originated from an academic competition for digital steganography, and comprises a large collection of greyscale photographs covering a wide variety of subjects and scenes. The algorithms are tested on selected images from the USC-SIPI dataset~\citep{2006_USC_SIPI}. All the images are resized to a resolution of $256 \times 256$ pixels via Lanczos resampling~\citep{1979_Lanczos}. The border pixels along with half of the rest of the pixels are designated as the context. Accordingly, the number of query pixels equals $(254 \times 254)/2$. We display both distortion and capacity as divided by the number of query pixels.

\begin{figure}
\centering
\subfloat[Aeroplane]{%
\resizebox*{5.5cm}{!}{\includegraphics{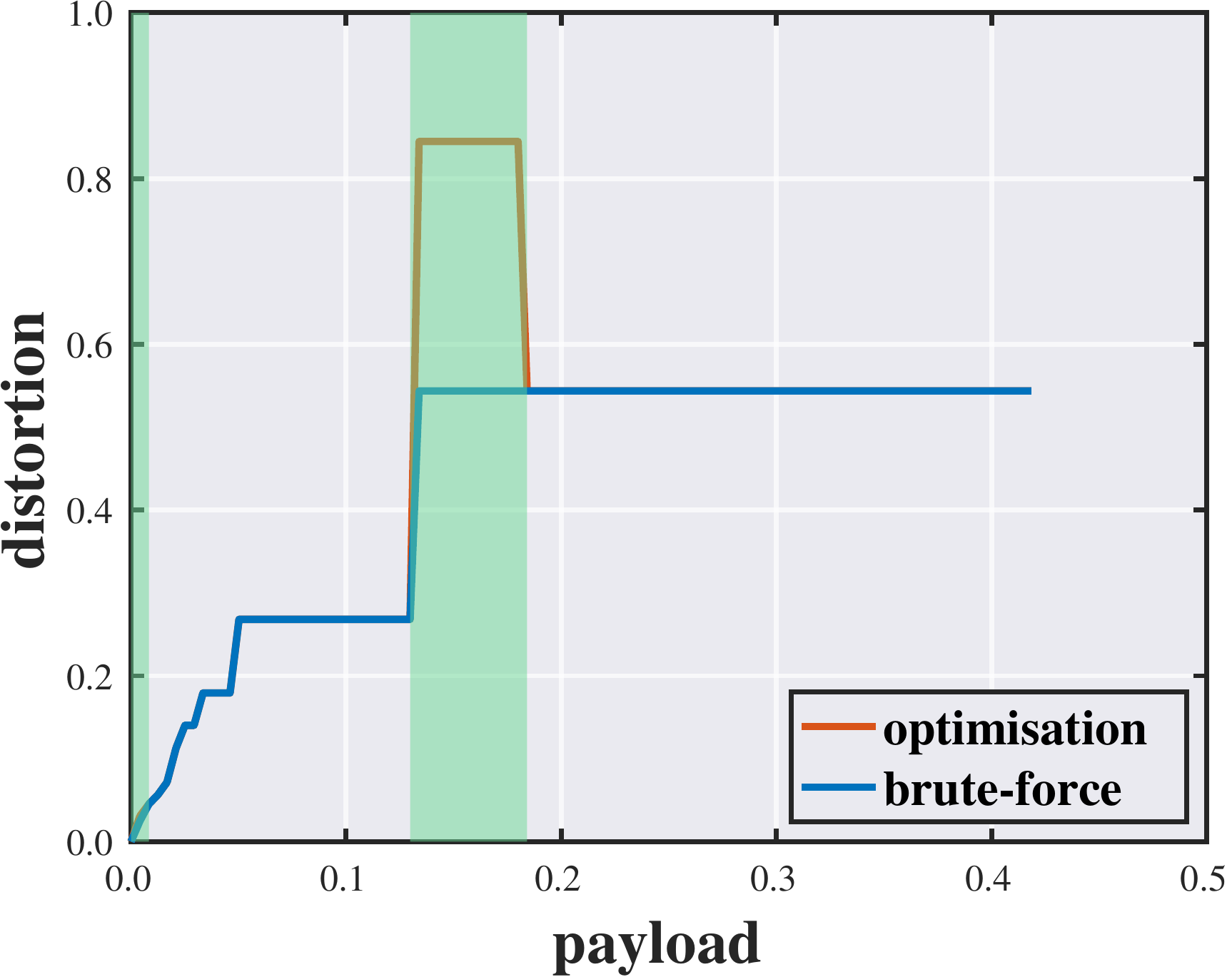}}}\hfil
\subfloat[Lena]{%
\resizebox*{5.5cm}{!}{\includegraphics{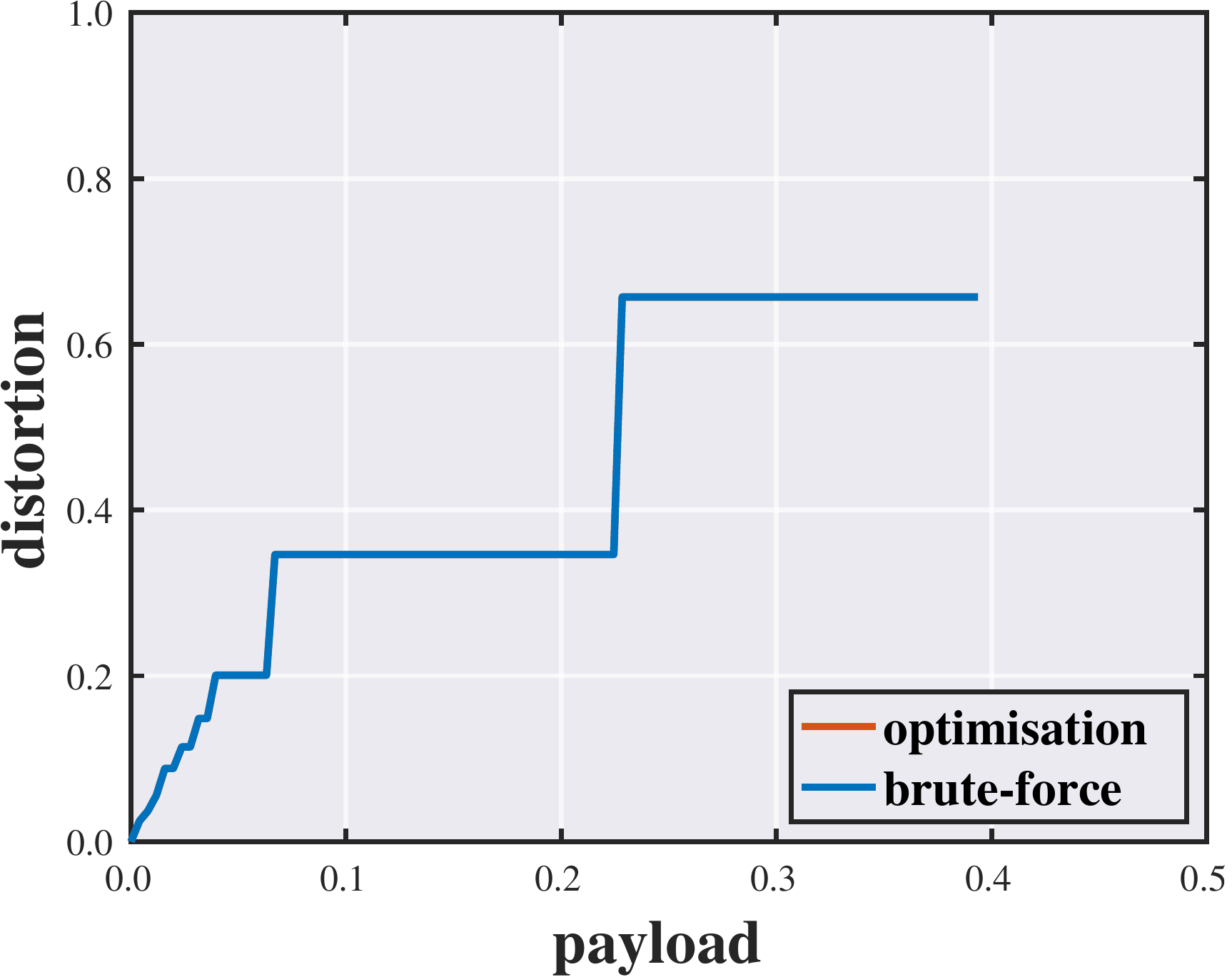}}}
\\
\subfloat[Mandrill]{%
\resizebox*{5.5cm}{!}{\includegraphics{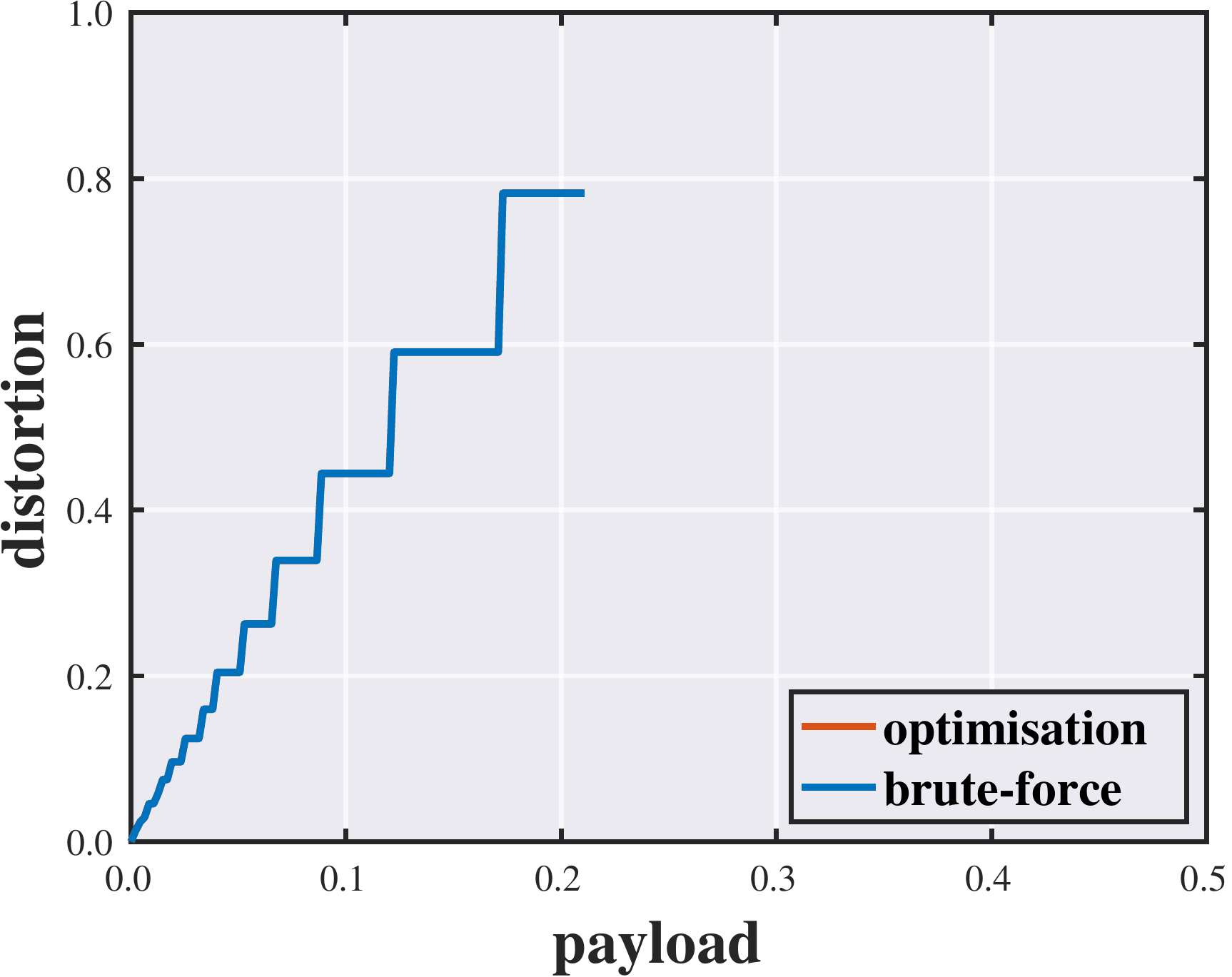}}}\hfil
\subfloat[Peppers]{%
\resizebox*{5.5cm}{!}{\includegraphics{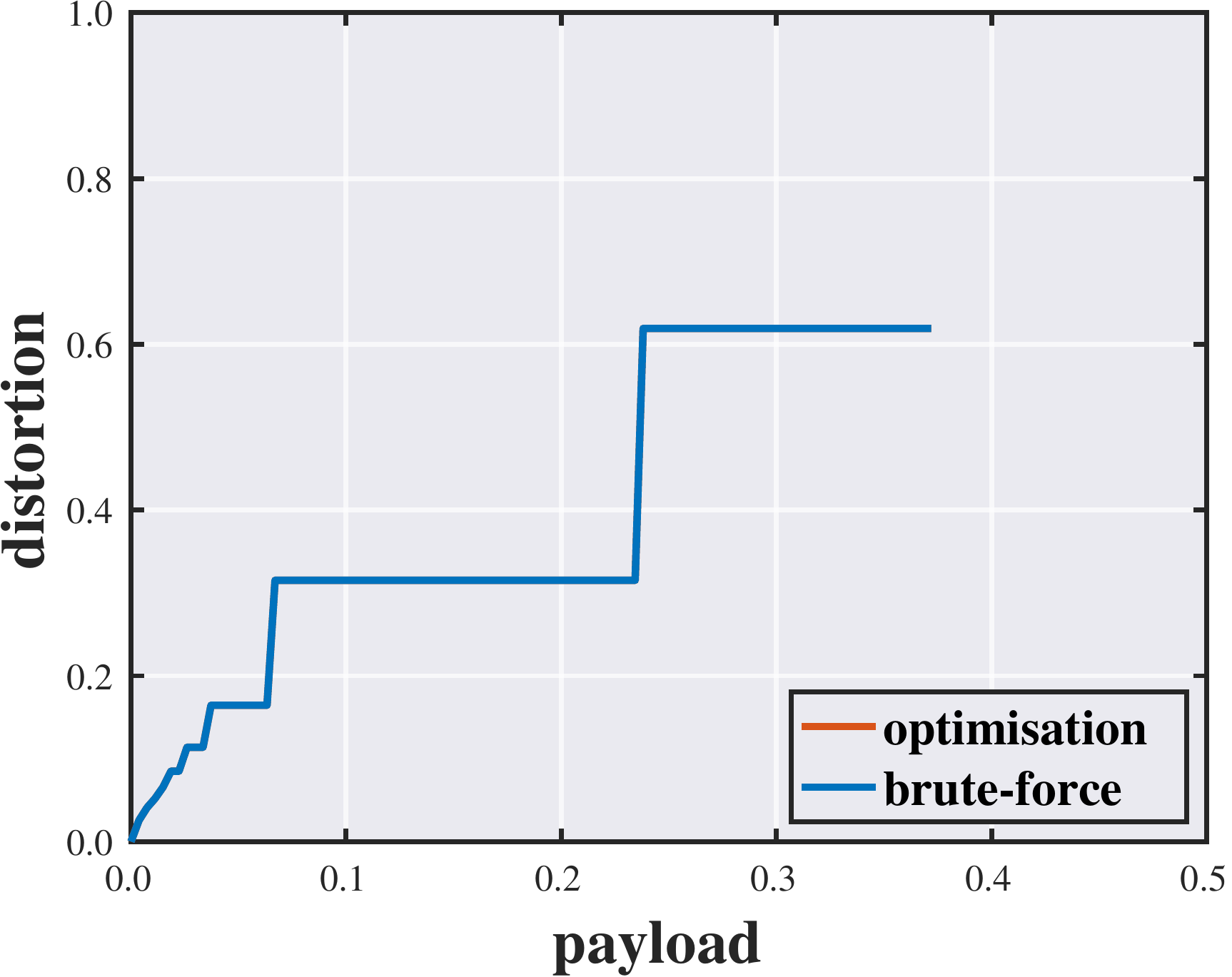}}}
\caption{Payload\textendash distortion curves for optimality analysis against brute-force search with highlighted discrepancies ($\vartheta=1$).} \label{fig:optim_analysis1}
\centering
\subfloat[Aeroplane]{%
\resizebox*{5.5cm}{!}{\includegraphics{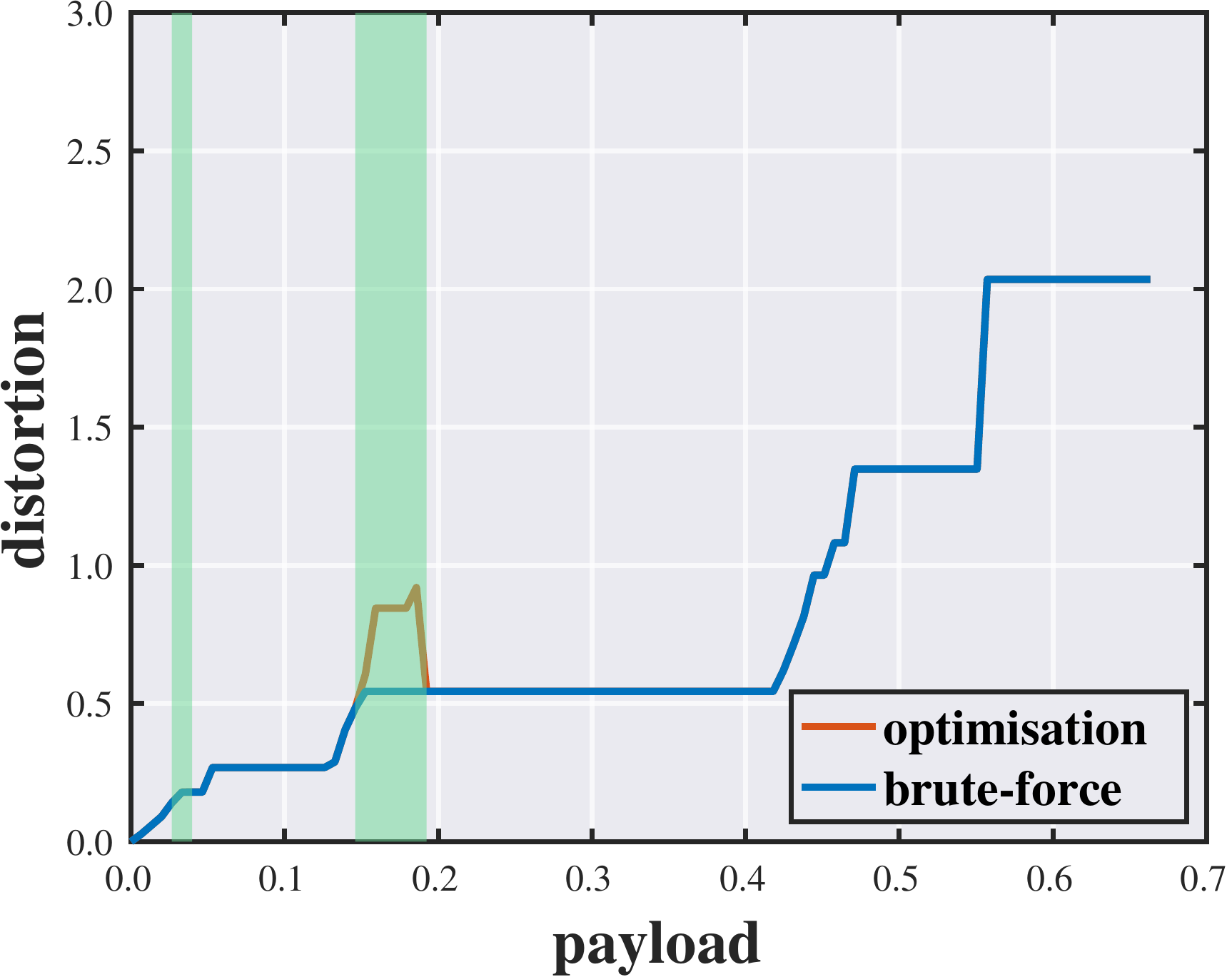}}}\hfil
\subfloat[Lena]{%
\resizebox*{5.5cm}{!}{\includegraphics{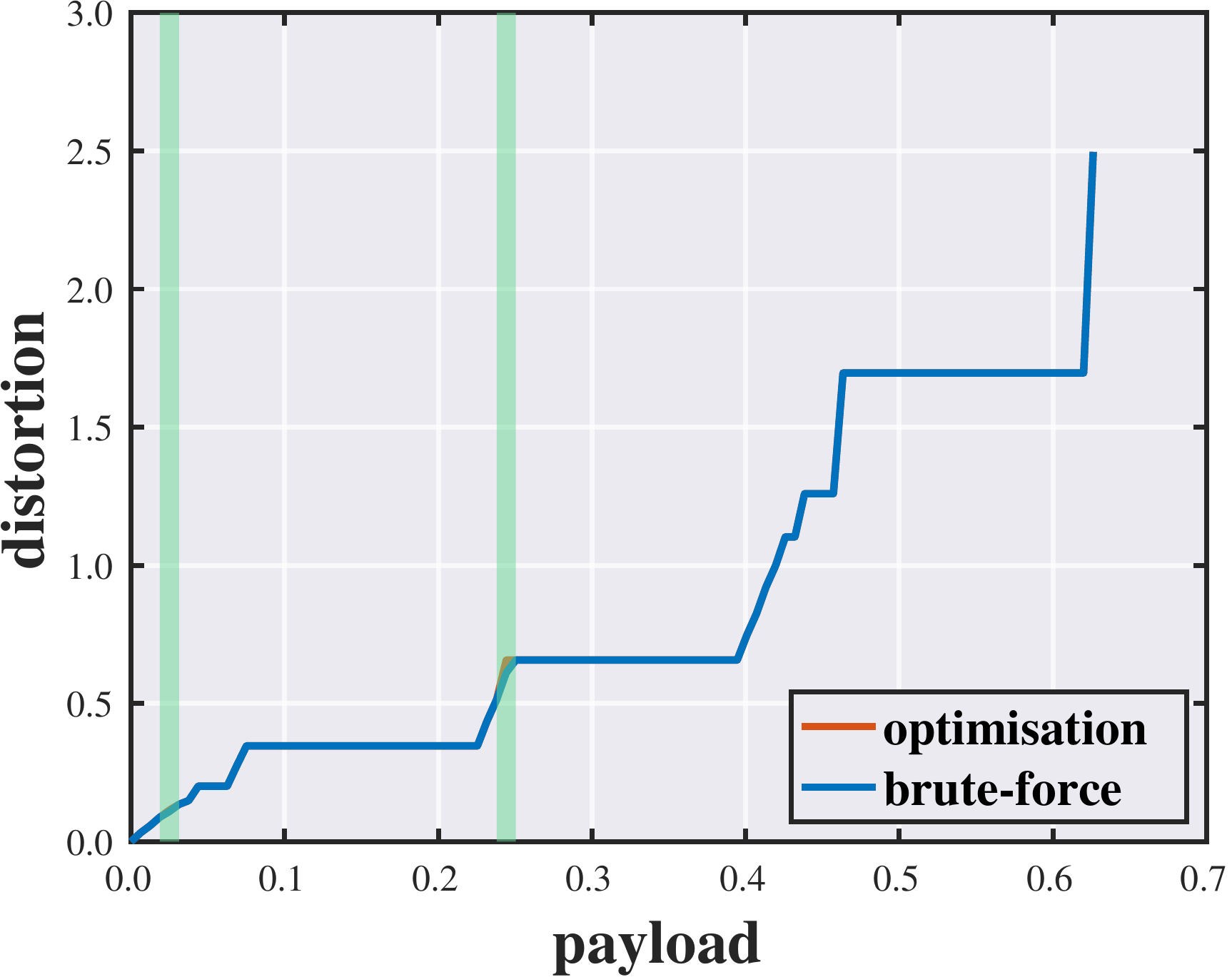}}}
\\
\subfloat[Mandrill]{%
\resizebox*{5.5cm}{!}{\includegraphics{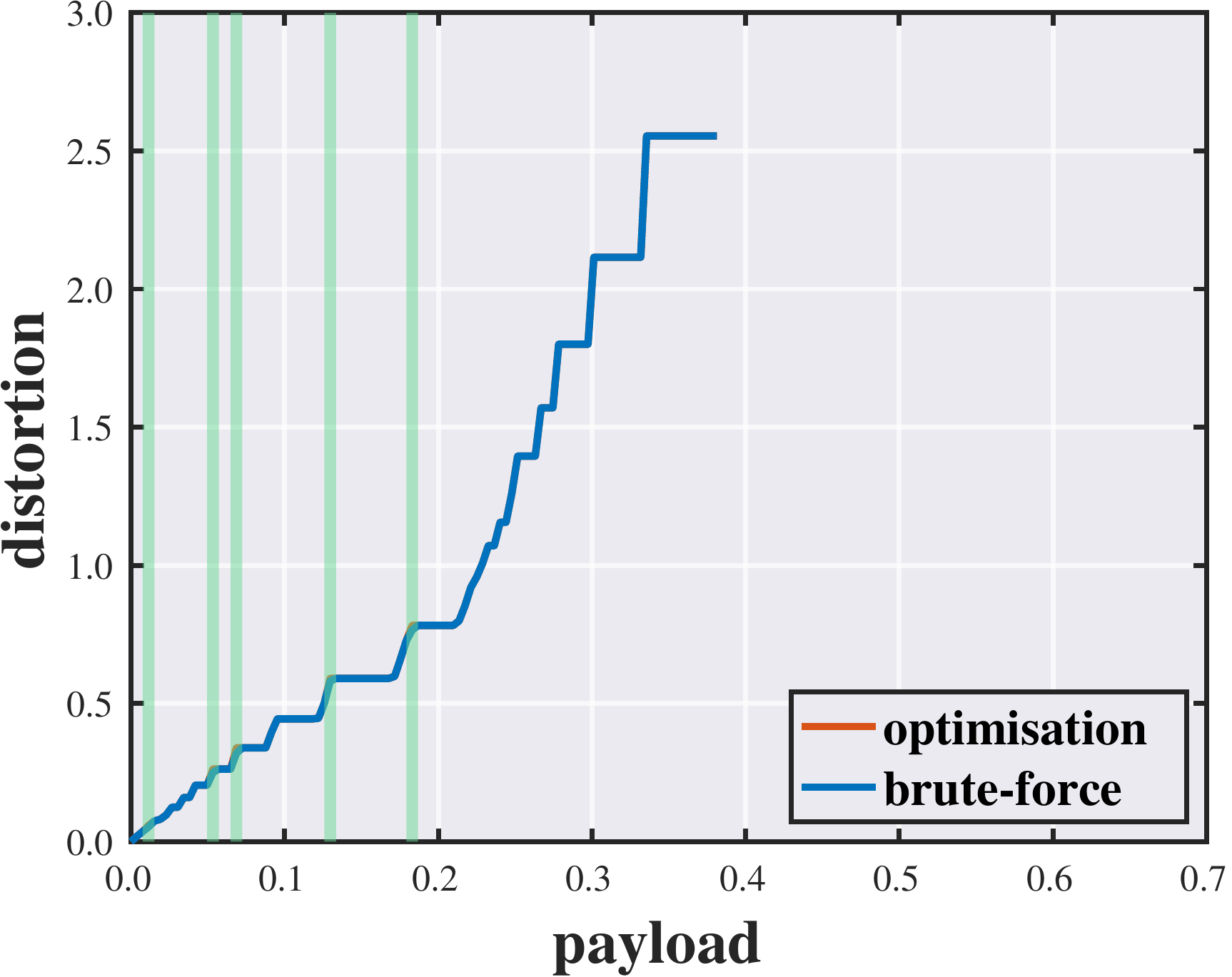}}}\hfil
\subfloat[Peppers]{%
\resizebox*{5.5cm}{!}{\includegraphics{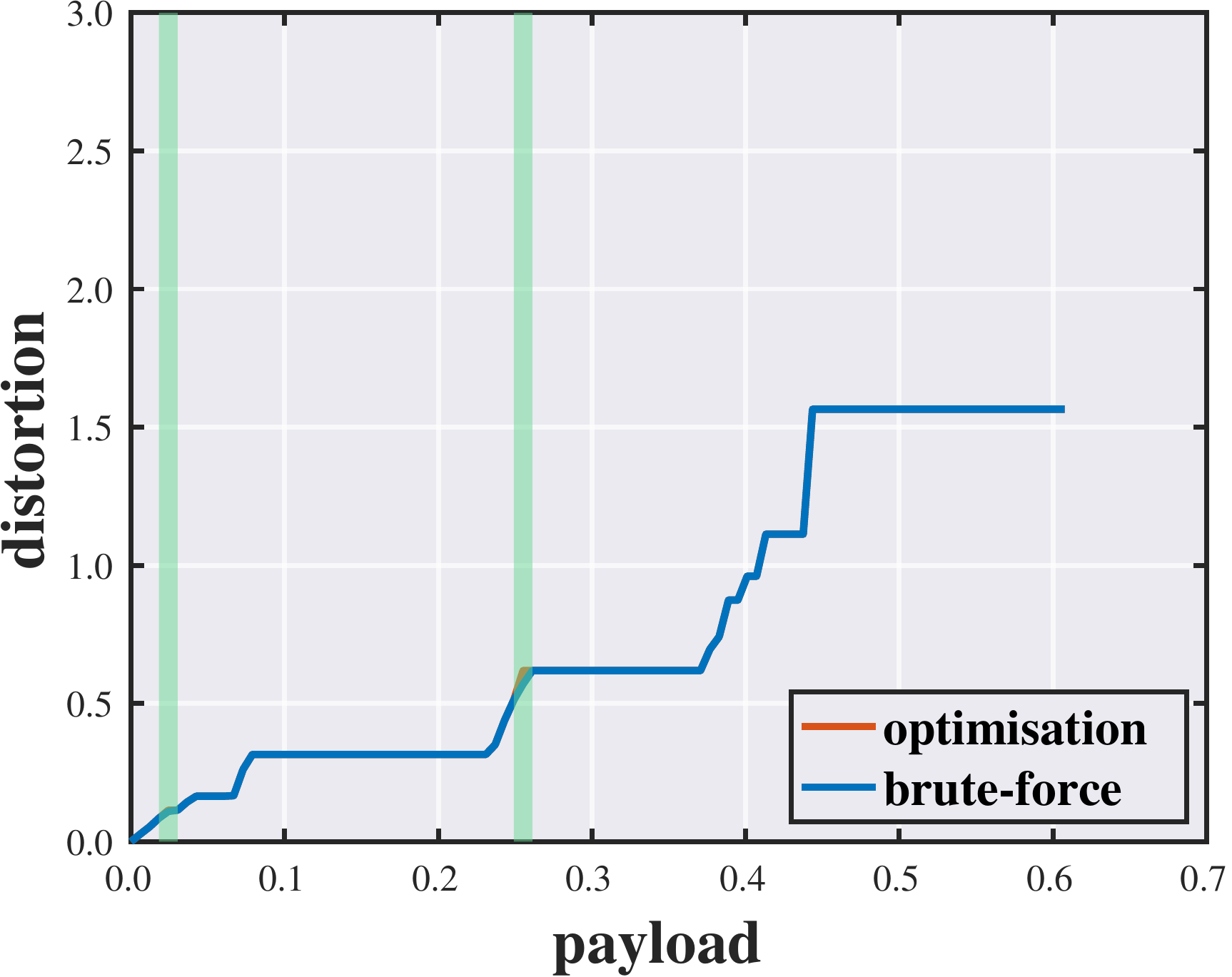}}}
\caption{Payload\textendash distortion curves for optimality analysis against brute-force search with highlighted discrepancies ($\vartheta=2$).} \label{fig:optim_analysis2}
\end{figure}

\begin{figure}
\centering
\subfloat[Aeroplane]{%
\resizebox*{5.5cm}{!}{\includegraphics{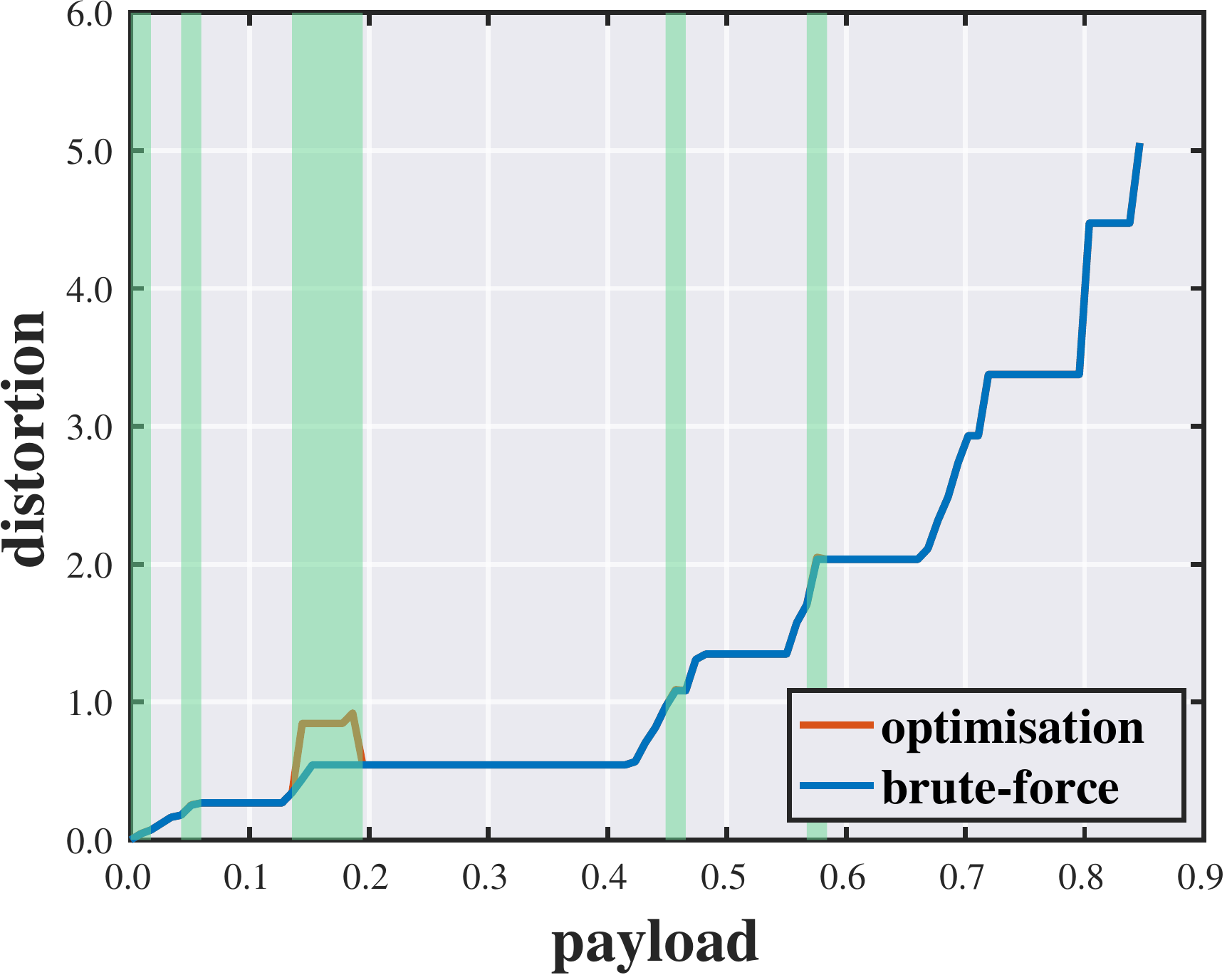}}}\hfil
\subfloat[Lena]{%
\resizebox*{5.5cm}{!}{\includegraphics{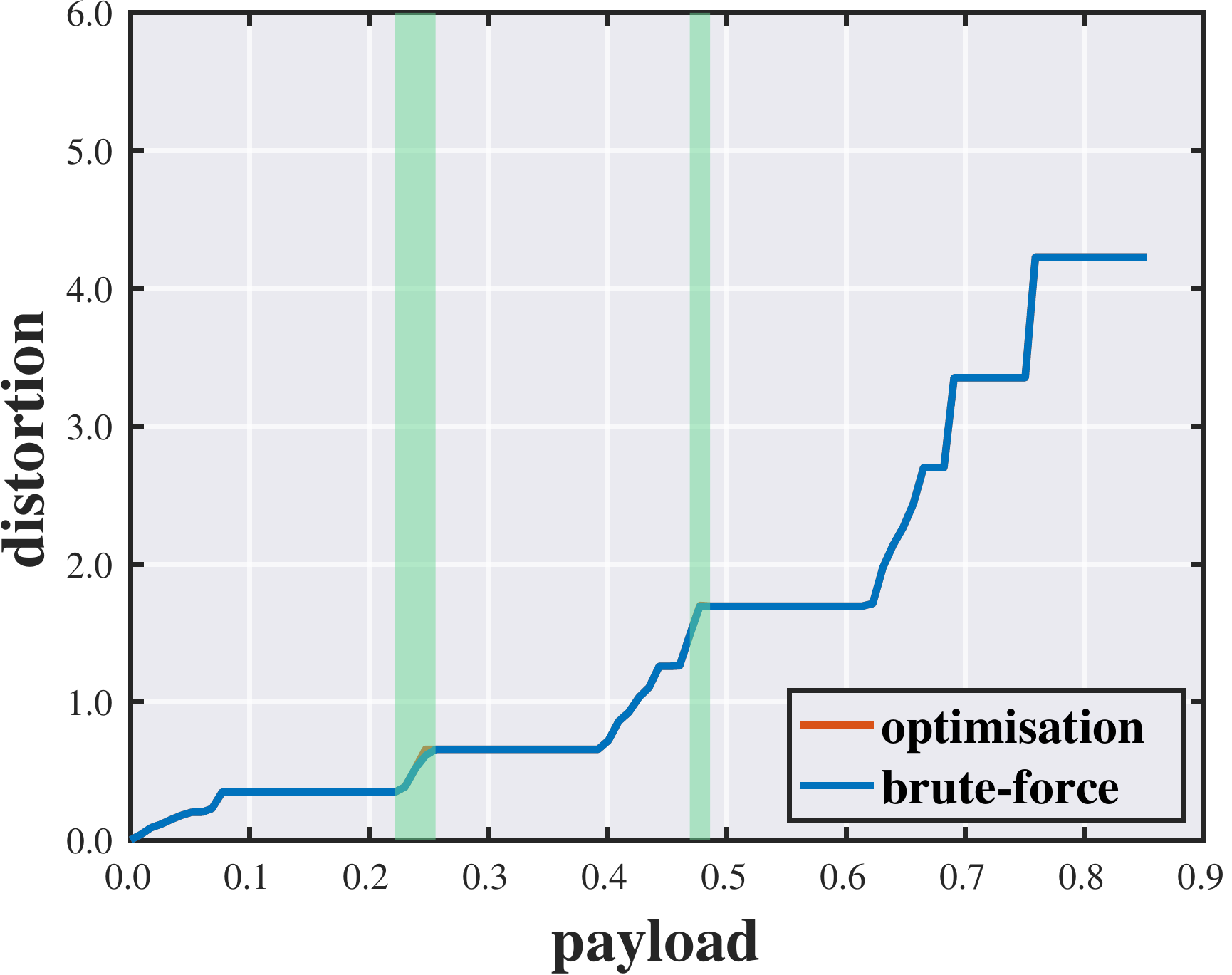}}}
\\
\subfloat[Mandrill]{%
\resizebox*{5.5cm}{!}{\includegraphics{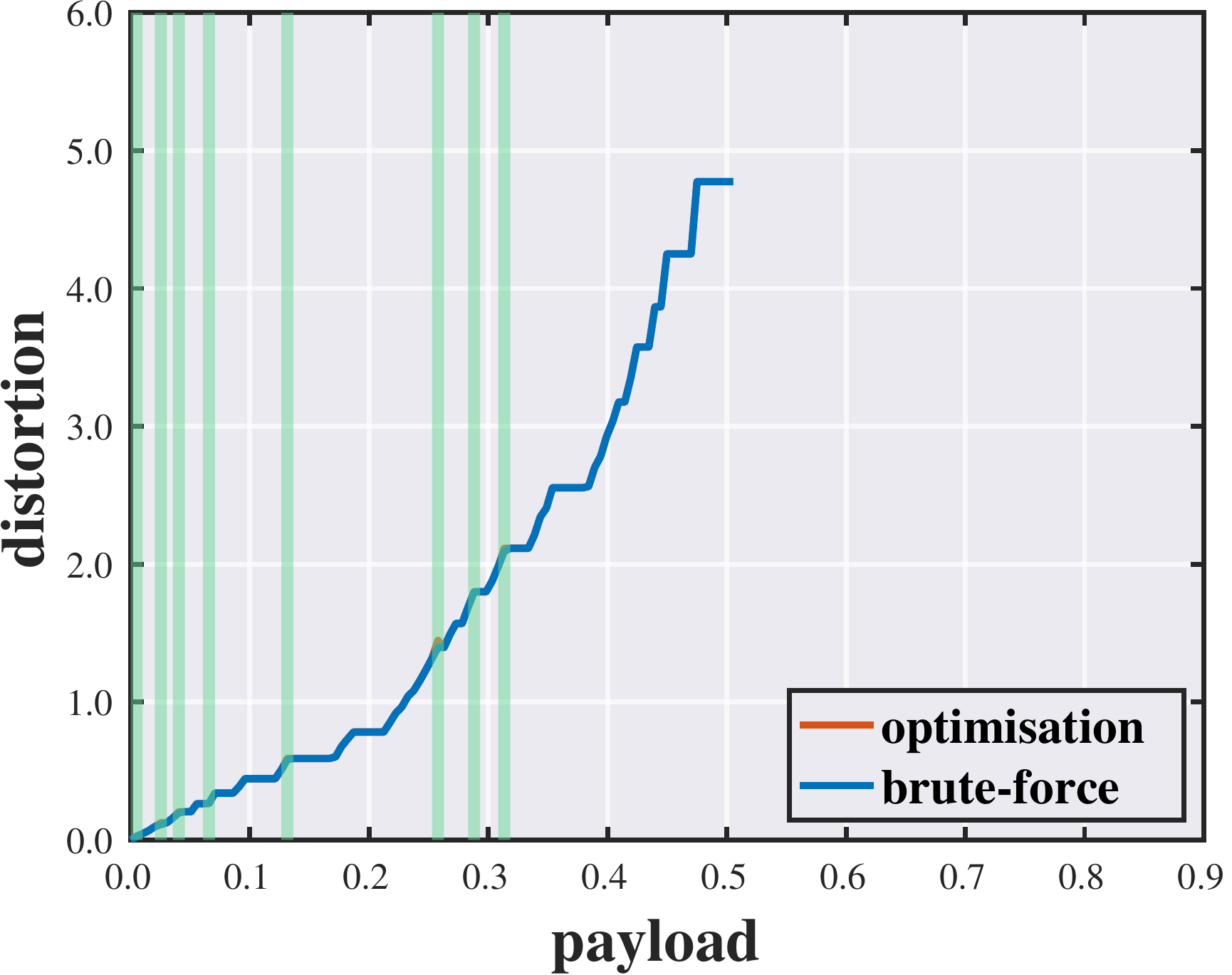}}}\hfil
\subfloat[Peppers]{%
\resizebox*{5.5cm}{!}{\includegraphics{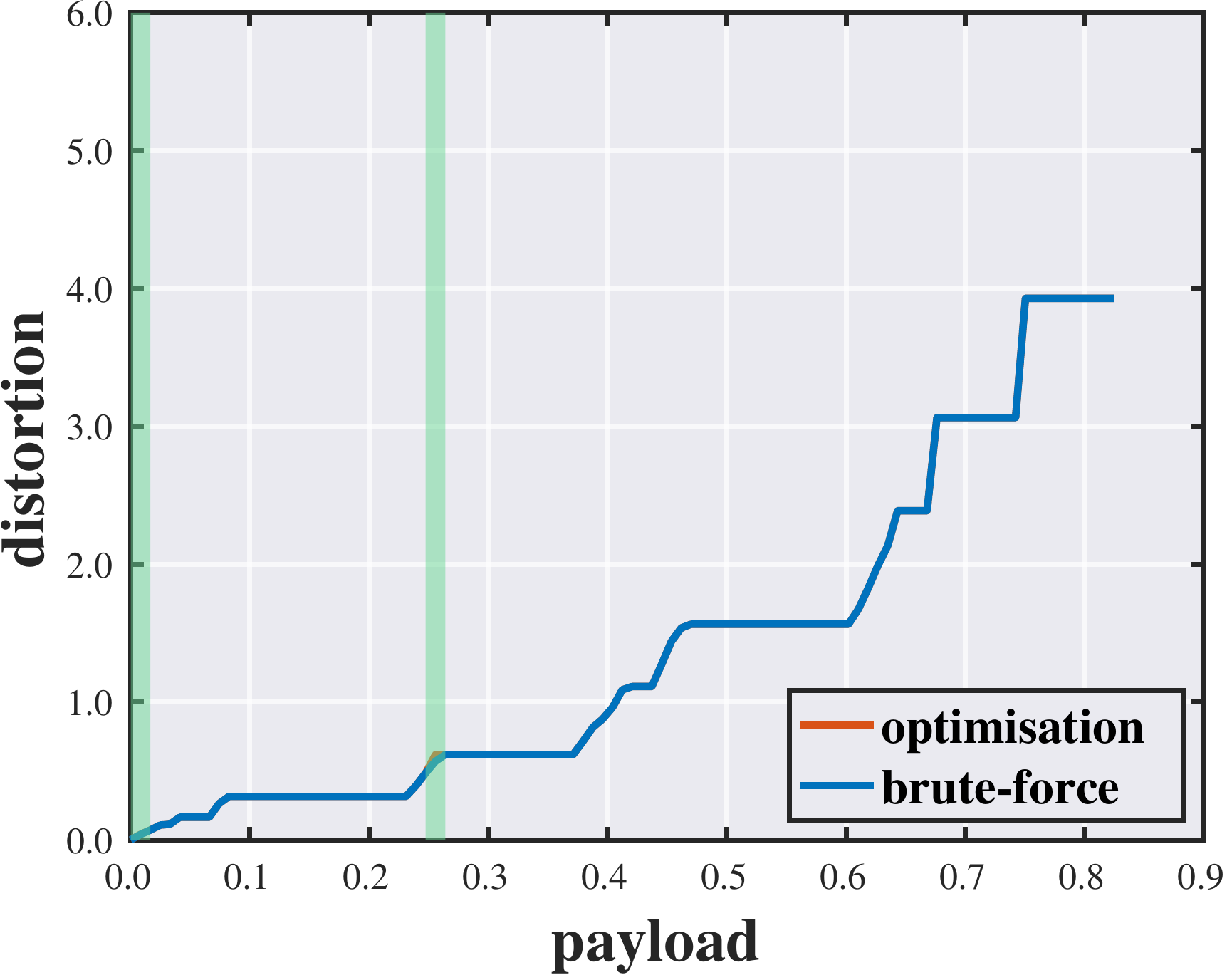}}}
\caption{Payload\textendash distortion curves for optimality analysis against brute-force search with highlighted discrepancies ($\vartheta=3$).} \label{fig:optim_analysis3}
\centering
\subfloat[Aeroplane]{%
\resizebox*{5.5cm}{!}{\includegraphics{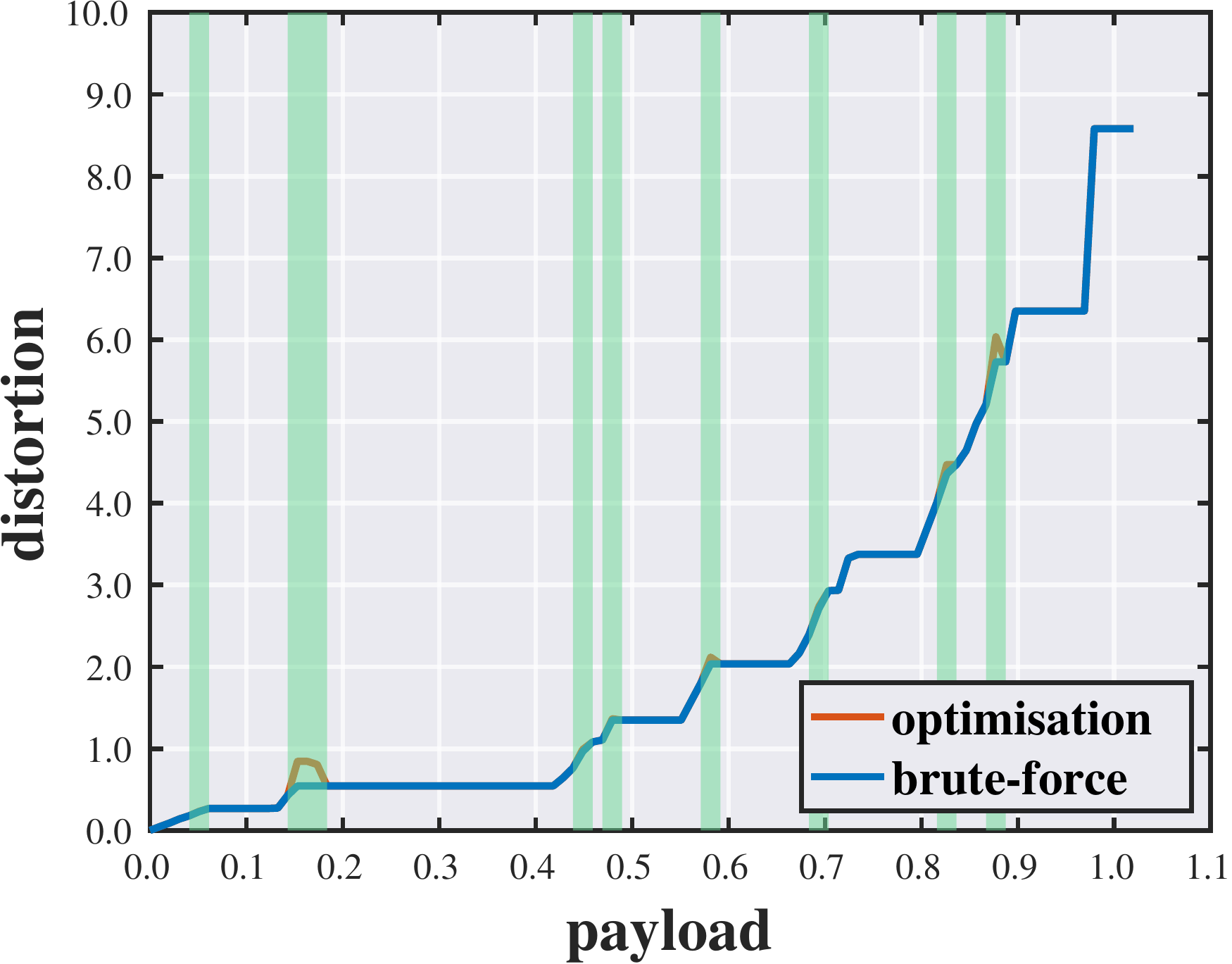}}}\hfil
\subfloat[Lena]{%
\resizebox*{5.5cm}{!}{\includegraphics{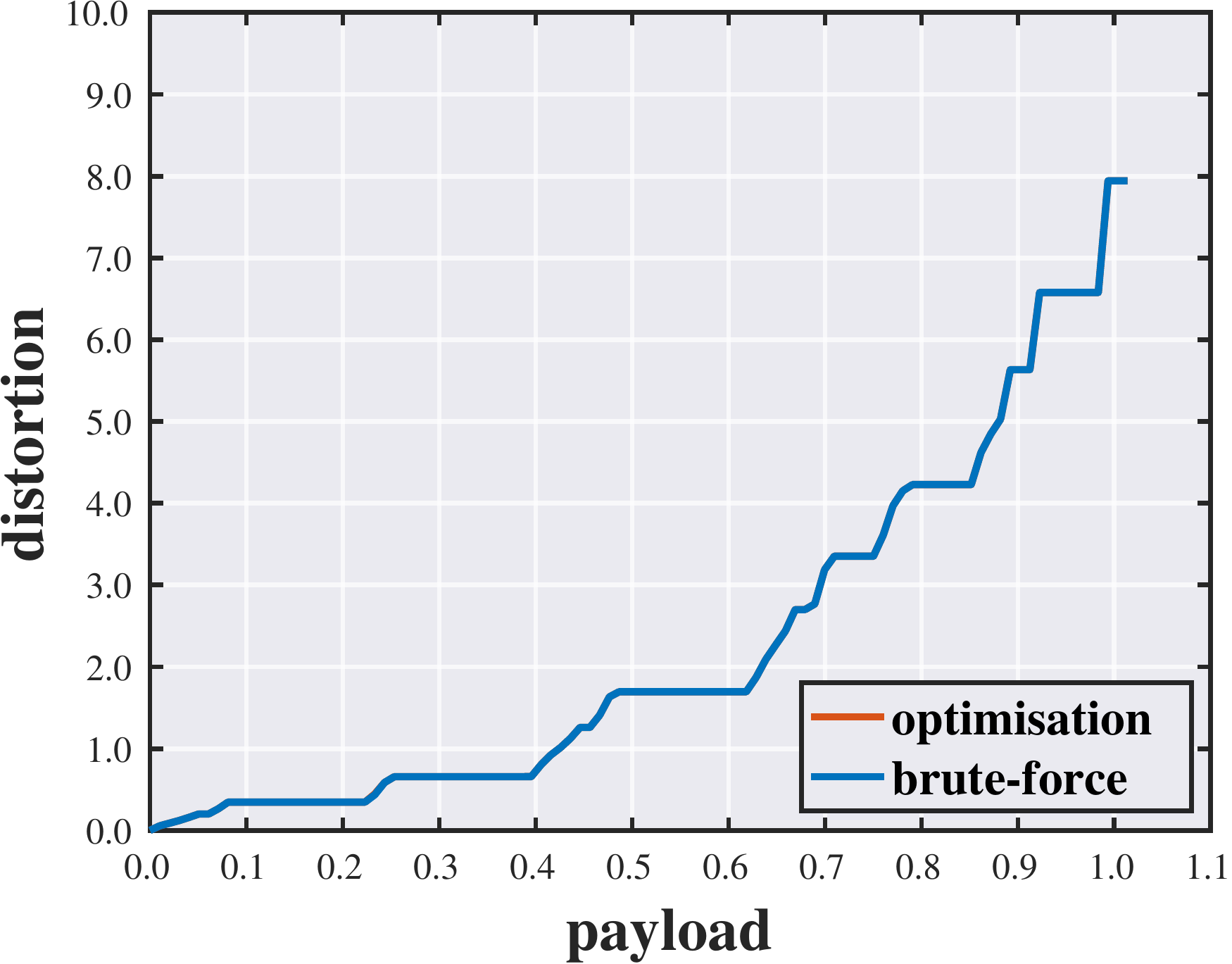}}}
\\
\subfloat[Mandrill]{%
\resizebox*{5.5cm}{!}{\includegraphics{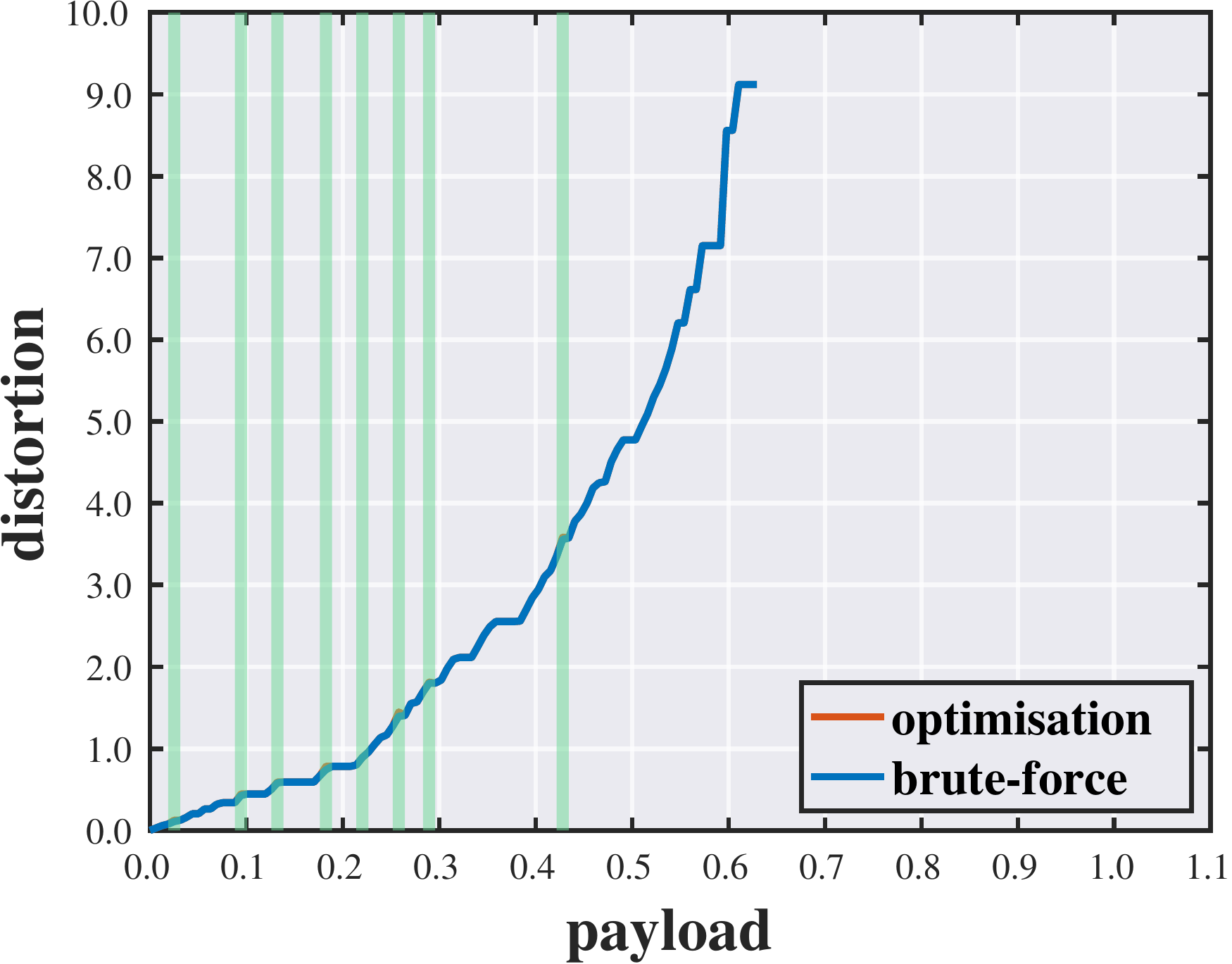}}}\hfil
\subfloat[Peppers]{%
\resizebox*{5.5cm}{!}{\includegraphics{Figures/exp/optim_10004_n55_theta3.pdf}}}
\caption{Payload\textendash distortion curves for optimality analysis against brute-force search with highlighted discrepancies ($\vartheta=4$).} \label{fig:optim_analysis4}
\end{figure}

Figure~\ref{fig:img_hist} shows the absolute error distribution for each test image. It is observed that most of the error values are below around $30$ to $50$, depending on the image. We set $n=55$ conservatively in the sense that nearly every value of non-zero occurrence is included. We implement the algorithms with respect to different quota settings ($\vartheta = 1, 2, 3, 4$). Figures~\ref{fig:optim_analysis1} to~\ref{fig:optim_analysis4} show performance evaluations of the proposed optimisation algorithm. Each point of the curve indicates the minimum distortion of a solution under a specific capacity constraint. In the vast majority of cases, the solutions found by the proposed method are identical to those given by the brute-force method. When failing to find the optimal solutions, the objective values reached are within a small distance from the optimal ones. Hence, even though optimal solutions cannot always be guaranteed, the results suggest that the proposed method can attain near-optimal performance.

\section{Conclusion}\label{sec:conclusion}
This paper studies a mathematical optimisation problem applied to reversible steganography. We formulate automatic coding in prediction error modulation as a nonlinear discrete optimisation problem. The objective is to minimise distortion under a constraint on capacity. We discuss the complexity of a brute-force search algorithm and the linearisation techniques for the logarithmic capacity constraint and the quadratic distortion objective. The problem is transformed into an iterative mixed-integer linear programming problem with binary-integer variables and slack variables. Our simulation results validate the near-optimality of the proposed algorithm.

\section*{Disclosure statement}
No potential conflict of interest is reported by the author.

\section*{Notes on contributors}
Ching-Chun Chang received his PhD in computer science from the University of Warwick, UK, in 2019. He participated in a short-term scientific mission supported by European Cooperation in Science and Technology Actions at the Faculty of Computer Science, Otto von Guericke University Magdeburg, Germany, in 2016. He was granted the Marie-Curie fellowship and participated in a research and innovation staff exchange scheme supported by Marie Sk\l{}odowska-Curie Actions at the Department of Electrical and Computer Engineering, New Jersey Institute of Technology, USA, in 2017. He was a visiting scholar at the School of Computer and Mathematics, Charles Sturt University, Australia, in 2018, and at the School of Information Technology, Deakin University, Australia, in 2019. He was a research fellow at the Department of Electronic Engineering, Tsinghua University, China, in 2020. His research interests include steganography, watermarking, forensics, biometrics, cybersecurity, applied cryptography, image processing, natural language processing, computer vision, computational linguistics, machine learning and artificial intelligence.

\bibliographystyle{apacite}
\bibliography{./Bib/myBib_abbrv}

\end{document}